\documentclass{article}

\usepackage[final,nonatbib]{neurips_2021}

\usepackage[utf8]{inputenc} 
\usepackage[T1]{fontenc}    
\usepackage{hyperref}       
\usepackage{url}            
\usepackage{booktabs}       
\usepackage{amsfonts}       
\usepackage{nicefrac}       
\usepackage{microtype}      
\usepackage{xcolor}         
\usepackage{graphicx}      
\usepackage{subcaption}
\usepackage{todonotes}

\usepackage[normalem]{ulem}
\usepackage{amsmath}
\usepackage{floatrow}
\usepackage{multirow}
\usepackage{numprint}
\usepackage{listings}       
\usepackage{xcolor}
\definecolor{codegreen}{rgb}{0,0.6,0}
\definecolor{codegray}{rgb}{0.5,0.5,0.5}
\definecolor{codepurple}{rgb}{0.58,0,0.82}
\definecolor{backcolour}{rgb}{0.95,0.95,0.96}
\usepackage[scaled=0.8]{FiraMono}
\lstdefinestyle{codestyle}{
    backgroundcolor=\color{backcolour},   
    commentstyle=\color{codegreen},
    keywordstyle=\color{magenta},
    numberstyle=\tiny\color{codegray},
    stringstyle=\color{codepurple},
    basicstyle=\ttfamily\fontsize{8}{9.5}\selectfont,
    breakatwhitespace=false,         
    breaklines=true,                 
    captionpos=b,                    
    keepspaces=true,                 
    numbersep=5pt,                  
    showspaces=false,                
    showstringspaces=false,
    showtabs=false,                  
    tabsize=2,
    language={python}
}
\title{
    CAPE: Encoding Relative Positions with \\
    Continuous Augmented Positional Embeddings
}

\author{%
    Tatiana Likhomanenko\thanks{Currently at Apple.} \\ 
    Facebook AI Research\\
    \url{tata.antares@gmail.com} \\
    \And
    Qiantong Xu \\
    Facebook AI Research \\
    \texttt{qiantong@fb.com} \\
    \And 
    Gabriel Synnaeve \\
    Facebook AI Research\\
    \texttt{gab@fb.com} \\
    \And
    Ronan Collobert \\
    Facebook AI Research\\
    \texttt{locronan@fb.com} \\
    \And 
    Alex Rogozhnikov \\
    Herophilus, Inc. \\
    \texttt{alex.rogozhnikov@yandex.ru} \\
}

\begin{document}

\maketitle

\begin{abstract}
Without positional information, attention-based Transformer neural networks are permutation-invariant. Absolute or relative positional embeddings are the most popular ways to feed Transformer models with positional information. Absolute positional embeddings are simple to implement, but suffer from generalization issues when evaluating on sequences longer than seen at training time. Relative positions are more robust to input length change, but are more complex to implement and yield inferior model throughput due to extra computational and memory costs. In this paper, we propose an augmentation-based approach (CAPE) for absolute positional embeddings, which keeps the advantages of both absolute (simplicity and speed) and relative positional embeddings (better generalization). In addition, our empirical evaluation on state-of-the-art models in machine translation, image and speech recognition demonstrates that CAPE leads to better generalization performance as well as increased stability with respect to training hyper-parameters.

\end{abstract}

\section{Introduction}
Transformers have been shown to be highly effective on problems involving sequential modeling, such as machine translation (MT)~\cite{vaswani2017attention} and natural language processing (NLP)~\cite{devlin2019bert,roller2020recipes,brown2020language}. 
Following its success on these tasks, the Transformer architecture raised immediate interest in other domains:
automatic speech recognition (ASR)~\cite{dong2018speech,gulati2020conformer}, music generation~\cite{huang2018music}, object detection~\cite{carion2020end}, and finally image recognition~\cite{dosovitskiy2020image,touvron2020training} and video understanding~\cite{bertasius2021space}.

Two major components of the Transformer are the attention mechanism~\cite{bahdanau2014neural,vaswani2017attention} and the positional encoding~\cite{vaswani2017attention,shaw2018self,huang2018music,dai2019transformer}. 
Without the latter, vanilla attention Transformers are invariant with respect to input tokens permutations (making ``cat eats fish'' and ``fish eats cat'' identical to the model). In the original Transformer publication,
 sinusoidal positional encoding was introduced~\cite{vaswani2017attention}. Token positions were encoded in an absolute manner, which was sufficient to achieve state-of-the-art performance in numerous tasks. However, performance issues were later observed when dealing with sequences of length not seen at training time~\cite{huang2018music,dai2019transformer,zhou2019improving,mohamed2019transformers,rosendahl2019analysis,dufter2021position}. For most applications relative positions between tokens are more relevant than absolute ones.
A number of approaches were thus proposed to encode relative positions in an explicit form~\cite{shaw2018self,dai2019transformer,huang2018music,rosendahl2019analysis,huang2020improve}, leading to improvements in modeling long sequences. 
However, all these approaches focus on modifying the attention mechanism and suffer from additional computational and memory costs~\cite{zhang2020pushing}. 
Relative positional encoding is also notably hard to implement efficiently for multidimensional case, and recent advances in Transformer models for computer vision~\cite{carion2020end,dosovitskiy2020image,touvron2020training} still rely on learnable absolute positional encoding.

Instead of changing the Transformer attention mechanism, we propose to improve absolute sinusoidal positional encodings in two ways:
a) instead of discrete positions, rely on continuous ones, which better match the continuous nature of image, sound or video data;
b) preserve some information about relative token positions, via a specific augmentation approach for positional embeddings during training. 
We empirically evaluate our approach, dubbed continuous augmented positional embedding (CAPE), with recent state-of-the-art models in several application domains. We study generalization properties and introduce unique features unlocked by CAPE. The main contributions of this work are: 
\begin{itemize}
    \item new augmented continuous positional embedding (CAPE), which encodes some relative position information in a computationally efficient way, and improves generalization performance compared to other positional embeddings across a variety of domains: machine translation, image and speech recognition;
    \item a single vision Transformer (UniViT) trained with CAPE on the mix of different resolutions: it outperforms each single-resolution baseline, generalizes better to unseen resolutions and can naturally process images of any size;
    \item new CAPE-based adaptive training scheme for ASR that eliminates need for padding.
\end{itemize}

\section{Related Works}

Since the appearance of Transformers, many works have investigated ways to encode positional information. A detailed analysis of various positional embeddings is available for BERT architecture~\cite{wang2021position}, where authors empirically relate properties of positional embeddings to performance on downstream NLP tasks. A recent study~\cite{ke2020rethinking} (also focuses on BERT) highlights the negative impact of spurious correlations between word and positional embeddings, and proposes to explicitly disentangle the contribution of positional and content embeddings in the attention mechanism.
In contrast, our approach implicitly enforces this disentanglement by leveraging augmentation.

Systematic studies of positional embeddings in audio domain are scarce. Several ways to encode relative positions for Transformer-based speech recognition are compared in~\cite{wang2020transformer}. 
Experiments show that absolute sinusoidal positional embeddings work no better than stacking consecutive frames at each time position (a particular form of convolution).
We provide a more thorough evaluation of positional embeddings in ASR, over multiple datasets. We also show that embeddings obtained from a one-layer convolutional frontend benefits from adding positional information.

Transformers for computer vision applications are still in their early days, and most works rely on learnable absolute positional embeddings only \cite{dosovitskiy2020image,touvron2020training,bertasius2021space,arnab2021vivit}.
Several recent works complement the Transformer architecture with convolutional layers to induce a spacial relationship between tokens~\cite{graham2021levit,chu2021conditional,wu2021cvt}.
As we discuss later, this restricts flexibility compared to pure attention-based models. The work~\cite{graham2021levit} suggests injecting learnable attention biases as an alternative mechanism to positional encoding. Evaluation of several positional encodings and their corresponding generalization has been done in a study \cite{chu2021conditional} which is in line with our work. Convolutional elements were introduced in the Transformer architecture, leading to better generalization properties.
In contrast, our experiments demonstrate that generalization can be achieved without architecture modification or convolutional inductive bias. 
Concerning video understanding, an evaluation of the impact of positional encoding was done in~\cite{neimark2021video}: according to the results, 
positional encoding-free architectures perform best. Other work~\cite{bertasius2021space} reports that adding absolute positional embeddings improves models performance, but contribution of encoding space and time vary between datasets.

As a summary, many positional embeddings variants were previously introduced, often modality-specific. In our cross-modal study we focus on generalization properties of popular and widely used embeddings, and improve on absolute sinusoidal positional embedding, leading to a flexible Transformer architecture, with great generalization properties across a number of different tasks.

The closest idea to our work is augmentation of positions in Universal Transformer~\cite{dehghani2018universal} where \textit{discrete} global shifts are applied to synthetic tasks for encoder-decoder Transformer models. 
In our work, we introduce \textit{continuous} augmentations, not \textit{discrete}, which are more natural for continuous modalities like images and speech where CAPE benefits most and which are not discussed in~\cite{dehghani2018universal}. 
In addition to global shifts our augmentations include global scaling and local shifts, and we also synchronize augmentations between encoder and decoder (Section~\ref{sec:mt}).

\section{Theoretical Analysis of Absolute Sinusoidal Positional Embeddings}\label{sec:sinpos}

Originally absolute positional $K$-dimensional embeddings were introduced in~\cite{vaswani2017attention} as
\begin{equation}
 E_{2k}(n)     = \cos{\omega_k n} \qquad
 E_{2k + 1}(n) = \sin{\omega_k n} \qquad
 \omega_k = 10000^{-2k/K} \qquad n\in\mathbb{Z}^+
 \label{eq:vashwani}
\end{equation}

with $k=1,2,\dots, K/2$ enumerating components for a token at position $n$. 
For simplicity of analysis, we rewrite Eq.~(\ref{eq:vashwani}) as a complex-valued embeddings\footnote{Work~\cite{Wang2020Encoding} introduces general complex-valued embeddings as continuous word functions over a position.} with half the number of components:
$$
    \{\mathbf{E}(n)\}_k = E_{k}(n) = e^{i \omega_k n}
$$

This definition can be rewritten in a recursive manner, by introducing a unitary operator $S$:
\begin{equation}
\label{eq-unitary-operator}
     \mathbf{E}(n+1) = S\,\mathbf{E}(n)\,,
     \qquad \textrm{with} \qquad \{S\,\mathbf{X}\}_k = X_k e^{i \omega_k}
\end{equation}

Therefore, the embedding at position $n$ contains sufficient information to compute the embedding of the next or previous positions, as applying $S^m$ ($m \in \mathbb{Z}$) performs a relative shift: $\mathbf{E}(n+m) = S^m\,\mathbf{E}(n)$. 
Variation in $\omega_i$ ensures that no positions $<10^4$ are assigned similar embeddings.
Before introducing augmentations of positional embeddings, we revisit positions parametrizations for different modalities.

\paragraph{Positional encoding for text}
For natural language processing it is common to split text into words, letters, syllables and other sub-units.
Original absolute sinusoidal positional embeddings enumerate these sub-units by their ordinal number $n$, a common choice that we follow.

\paragraph{Positional encoding for images}
In a framework where patch and image sizes may vary, we find that enumerating patches is not appropriate, as positional embeddings may greatly differ for different scales of an image, leading to generalization issues. In that perspective, we consider scaled coordinates $x$ and $y$ that span interval $[-1, +1]$.\footnote{Authors of~\cite{carion2020end} also perform coordinates scaling but it spans interval $[0, 1]$.} While previous works~\cite{dosovitskiy2020image,touvron2020training} relied on learnable absolute positional embedding, we introduce the following $K$-dimensional absolute sinusoidal positional embedding defined for each position $(x, y) \in \mathbb{R}^2$:
\begin{equation}\label{eq:sin2d1}
E_{2k}(x, y)   = \cos \pi (w_{k,x} x + w_{k,y} y) \qquad
E_{2k+1}(x, y) = \sin \pi (w_{k,x} x + w_{k,y} y) 
\end{equation}
\begin{equation}\label{eq:sin2d2}
w_{k,x} = 10^{2k/K} \cos{k} \qquad
w_{k,y} = 10^{2k/K} \sin{k}
\end{equation}
Following Eq.~(\ref{eq-unitary-operator}), this corresponds to introducing two commuting unitary operators $S_x$ and $S_y$, for each unit shift in either direction on the plane. 
The choice of $w_{k,x}$ and $w_{k,y}$ is kept simple and deterministic, while giving no preference to any direction on the plane and providing angle of ``hatching'' (all components have different angle and angles uniformly cover possible directions), see Figure~\ref{fig:emb_visualization}. 
Furthermore, embeddings defined by Eq.~(\ref{eq:sin2d1}-\ref{eq:sin2d2}) allow attending to a specific small region around a point on the image without emphasizing points with only same $x$ or only same~$y$ while we expect this artifact to contribute in case of concatenation of 1D embeddings for each axis~\cite{chu2021conditional,dosovitskiy2020image,bello2019attention,parmar2018image}.

\paragraph{Positional encoding for sound}
We propose to tie positional embeddings to timestamps in seconds. The embedding for a frame centered at $t$ seconds is given by:
$$
    E_{2k}(t) = \cos{\omega_k t}
    \qquad E_{2k + 1}(t) = \sin{\omega_k t}
    \qquad \omega_k = 30 \times 10000^{-2k/K}
$$
The choice of $\omega_k$ corresponds to the scaled version of Eq.~(\ref{eq:vashwani}) and ensures that queries with 30ms specificity are possible even with minutes-long audio fragments.

Two main factors contribute to the choice of base in the exponentiation of $w_k$ and $w_{k,x}$, $w_{k,y}$ frequencies: i) how precise the location should be for encoding (e.g. there is no need to encode position in audio much more precisely than one syllable takes); ii) longest reasonable ``length'' that is expected in the data. CAPE could be parametrized by precision and ``length'' parameters, but we believe that for all practical cases provided choices are reasonable.

\section{Continuous Augmented Positional Embeddings (CAPE)}

\begin{figure}[t!]
  \centering
  \includegraphics[width=0.75\textwidth]{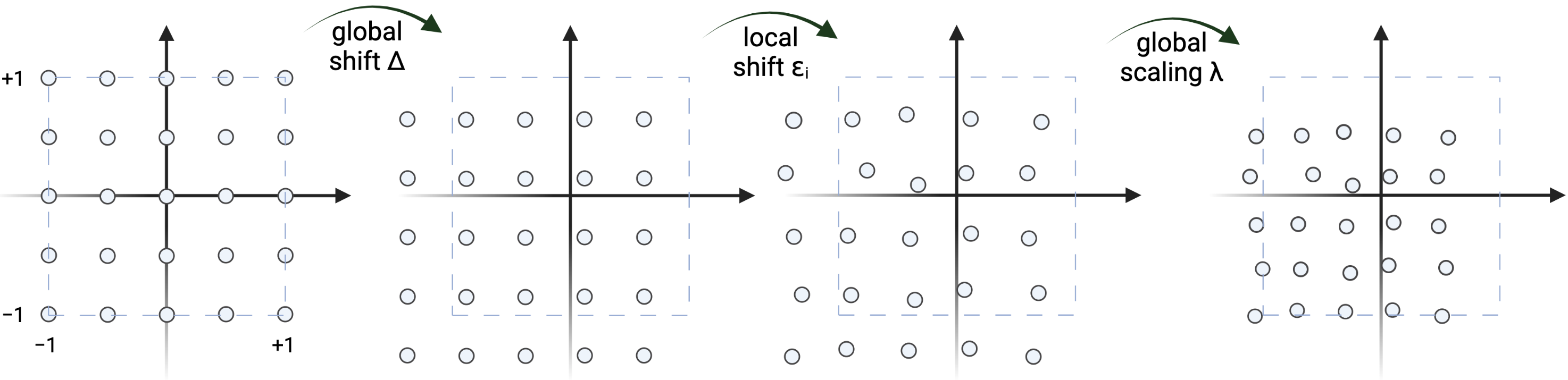}
  \caption{Example of CAPE's transformations for an image: patches positions are scaled to $[-1, 1]\times [-1, 1]$; random global shift, local shift, and global scaling are then applied to the grid of positions.
  \label{fig:cape}
  }
\end{figure}

Absolute sinusoidal positional embeddings lack regularization, leading to ``in-domain'' generalization issues as the model may start learning spurious correlations. For applications where inference input sizes may be significantly different than the training ones, ``out-of-domain'' issues may also arise, as positions rarely observed at training may lead to improper inference predictions.
To ensure that the model does not learn spurious correlations between content and position, we introduce the following augmentations for absolute sinusoidal positional embeddings at training time, see Figure~\ref{fig:cape}.

\paragraph{Global shift} 
Transform every embedding in a sequence using $S^\Delta$ operator with a global random shift from uniform zero-mean distribution $\Delta \sim \mathcal{U}(-\Delta_{max}, \Delta_{max})$:
$$
\mathbf{E}'(n) = S^\Delta \mathbf{E}(n) \qquad
\{S^\Delta\,\mathbf{X}\}_k = X_k e^{i \omega_k \Delta } \qquad \Delta \in \mathbb{R}
$$
This modification hides the absolute positional information, but relative relations, see Eq.~(\ref{eq-unitary-operator}), between embeddings still hold. 
This transformation can be rewritten as augmenting positions by a random shift before encoding with $\sin$ and $\cos$:
\begin{equation}
    n'_i \leftarrow n_i + \Delta \qquad \quad
    x'_i \leftarrow x_i + \Delta_x, \;\;
    y'_i \leftarrow y_i + \Delta_y \qquad \qquad 
    t'_i \leftarrow t_i + \Delta
    \label{cape-global-shift}
\end{equation}

\paragraph{Local shift}
To further increase augmentation and prevent capturing spontaneous correlations, 
we additionally introduce local shifts from uniform zero-mean distribution $\epsilon_i \sim \mathcal{U}(-\epsilon_{max}, \epsilon_{max})$
\begin{equation}
    n'_i \leftarrow n_i + \epsilon_i        \qquad \quad
    x'_i \leftarrow x_i + \epsilon_{x,i},   \;\;
    y'_i \leftarrow y_i + \epsilon_{y,i}    \qquad \qquad 
    t'_i \leftarrow t_i + \epsilon_i
    \label{cape-local-shift}
\end{equation}
\paragraph{Global scaling} To prevent distances memorization, we also introduce random global scale $\lambda$ from $\log \lambda \sim \mathcal{U}(- \log \lambda_{max}, \, \log \lambda_{max})$
\begin{equation}
    n'_i \leftarrow \lambda n_i     \qquad \qquad \quad
    x'_i \leftarrow \lambda x_i,    \;\;          \quad
    y'_i \leftarrow \lambda y_i     \qquad \qquad \quad \quad
    t'_i \leftarrow \lambda t_i                   \quad
    \label{cape-scale}
\end{equation}

At training time, computing our continuous augmented positional embedding is performed through four steps: i)~mean-normalization of positions (extract mean of sequence positions), ii)~global shift Eq.~(\ref{cape-global-shift}), iii)~local shift Eq.~(\ref{cape-local-shift}), and iv)~global scaling Eq.~(\ref{cape-scale}).
At inference time, only mean-normalization of positions is performed. The reference implementation can be found in Appendix~\ref{app:impl}.

\section{Experiments}

\subsection{Image Recognition}

We evaluate CAPE embedding empirically with a recently proposed Vision Transformer (ViT)~\cite{dosovitskiy2020image,touvron2020training} for image recognition. 
These works rely on learnable absolute positional embedding (\textit{abspos}) for both class token and patches, and train ViT models on $224^2$ images\footnote{In the following, we denote the size and resolution $N\times N$ as $N^2$.} with $16^2$ patches. 
To further improve model quality, \cite{touvron2020training} performs fine-tuning on images of higher resolution $384^2$. The grid of positional embeddings is then upsampled.

\paragraph{Data and ViT Models}
All experiments are performed on the ImageNet~\cite{deng2009imagenet,russakovsky2015imagenet} dataset. 
We report top-1 and top-5 accuracies on ImageNet validation set and ImageNet-v2-\{a,b,c\}~\cite{recht2019imagenet} test sets. The same convolution-free architecture, identical to the one proposed by~\cite{dosovitskiy2020image} (ViT-B) and used by~\cite{touvron2020training} (referred as DeiT-B), is chosen for all experiments. 
A ViT-B/DeiT-B baseline is trained with \textit{abspos} on $224^2$ images, carefully following Section~6 from~\cite{touvron2020training}. 
The \textit{exact same training configuration} is used for models with other positional embeddings: \textit{only} positional embedding is changed. We evaluate both proposed absolute sinusoidal positional embedding (\textit{sinpos}), Eq.~(\ref{eq:sin2d1}-\ref{eq:sin2d2}), and CAPE ($\Delta_{max}=0.5$, $\epsilon_{max}=1/N$ and $\lambda_{max}=1.4$). 
As a control experiment we also train a model without any positional embedding (\textit{nopos}), which can be interpreted as a 'bag of \sout{words} patches', as no patch position information is available.
We also train models with different positional embeddings on either $160^2$ or $384^2$ images. The whole training configuration remains the same as for training on $224^2$ images, except for the positional embedding. 
All models trained on $224^2$ images we additionally fine-tune on images of higher resolution $384^2$, following~\cite{touvron2020training}. 

\paragraph{Evaluation}
To study generalization when image sizes vary, we evaluate all models on different resolutions ($160^2$, $224^2$, $384^2$ and $672^2$) by resizing all images in validation and test sets.  
When evaluating on resolutions different from the training one, bicubic interpolation is applied to \textit{abspos} embeddings\footnote{Interpolation is not applied to the class token embedding.} as was justified in~\cite{touvron2020training}. In contrast, \textit{sinpos} and CAPE approaches can ingest any image resolution, thanks to the continuous nature of their positional embeddings.

\subsubsection{Results}

\begin{figure}[t!]
  \centering
  \includegraphics[width=0.95\textwidth]{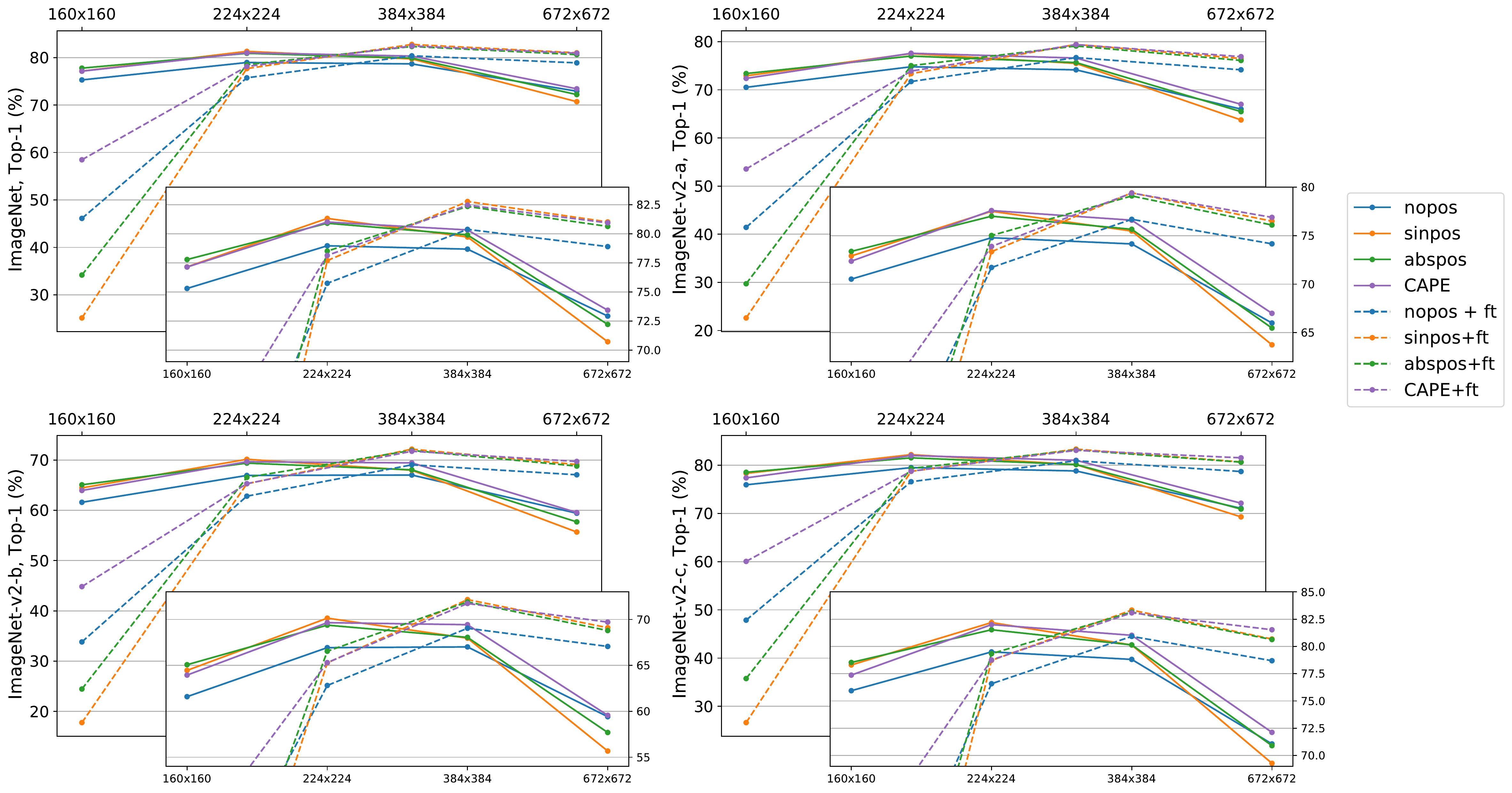}
  \caption{Top-1 accuracy on ImageNet and ImageNet-v2 for ViT models trained with different positional embeddings on $224^2$ resolution (solid) and further fine-tuned on $384^2$ (dashed, ``+ft''). Insets focus on higher accuracies.
  The full list of top-1 and top-5 accuracies can be found in Appendix~\ref{app:vision:tech}, Tables~\ref{tab:vision_all_top1} and~\ref{tab:vision_all_top5}. 
  \label{fig:vision:vit_main}
  }
\end{figure}

In Figure~\ref{fig:vision:vit_main} we compare generalization performance of models trained with different positional embeddings on $224^2$ images (solid). 
Both proposed \textit{sinpos} and CAPE approaches perform at least as well, if not better, than the \textit{abspos} approach on the same-as-train resolution. When performing inference on resolutions different than the training one, CAPE performs best, notably outperforming \textit{abspos} on high ($672^2$) and low ($160^2$) resolutions up to 25\% and 2\%, respectively.
On $160^2$ and $384^2$ resolutions CAPE trained on $224^2$ resolution performs similar to \textit{abspos} trained on corresponding $160^2$ or $384^2$ inference resolutions (the latter results being reported in Figure~\ref{fig:vision:mix_sizes}). This confirms good generalization properties of CAPE on image resolutions unseen at training time.

\textit{Abspos} fine-tuned on a higher resolution ($384^2$) improves in accuracy for both $384^2$ and $672^2$ resolutions, while degrading on lower ones (original $224^2$ and lowest $160^2$), as shown in Figure~\ref{fig:vision:vit_main} (dashed). 
\textit{Sinpos} and CAPE fine-tuned on $384^2$ resolution outperform \textit{abspos} by 0.3-0.4\%, thanks to a better fine-tuning starting point after resolution change. In that setting, \textit{sinpos} and CAPE keep better generalization performance on nearby resolutions ($224^2$ and $672^2$). While being seriously impacted on the $160^2$ resolution, CAPE still outperforms others by 12-25\%. Comparison of ViT models trained with either \textit{abspos} or CAPE on each particular resolution is shown in Figure~\ref{fig:vision:mix_sizes}: CAPE outperforms \textit{abspos} on a specific training resolution, and generalizes better to unseen resolutions.

\paragraph{No Positional Embedding}
With \textit{nopos} model, we confirm \cite{dosovitskiy2020image}'s observation that positional embedding does not have a critical importance for ImageNet classification (see Figure~\ref{fig:vision:vit_main}). \textit{Nopos} model's simplicity and generalization abilities make it a nice baseline for positional embedding study in computer vision. It has generalization accuracy similar to \textit{abspos} on low $160^2$ and high $672^2$ resolutions, while CAPE outperforms \textit{nopos} across the board. 
It is likely that \textit{abspos} suffers from the embedding interpolation step on extreme resolutions. In contrast, both \textit{sinpos} and CAPE have the advantage to naturally support variation in resolutions.

\subsubsection{UniViT: Training Universal Transformer on Different Resolutions}

\begin{figure}[t!]
  \centering
  \includegraphics[width=0.9\textwidth]{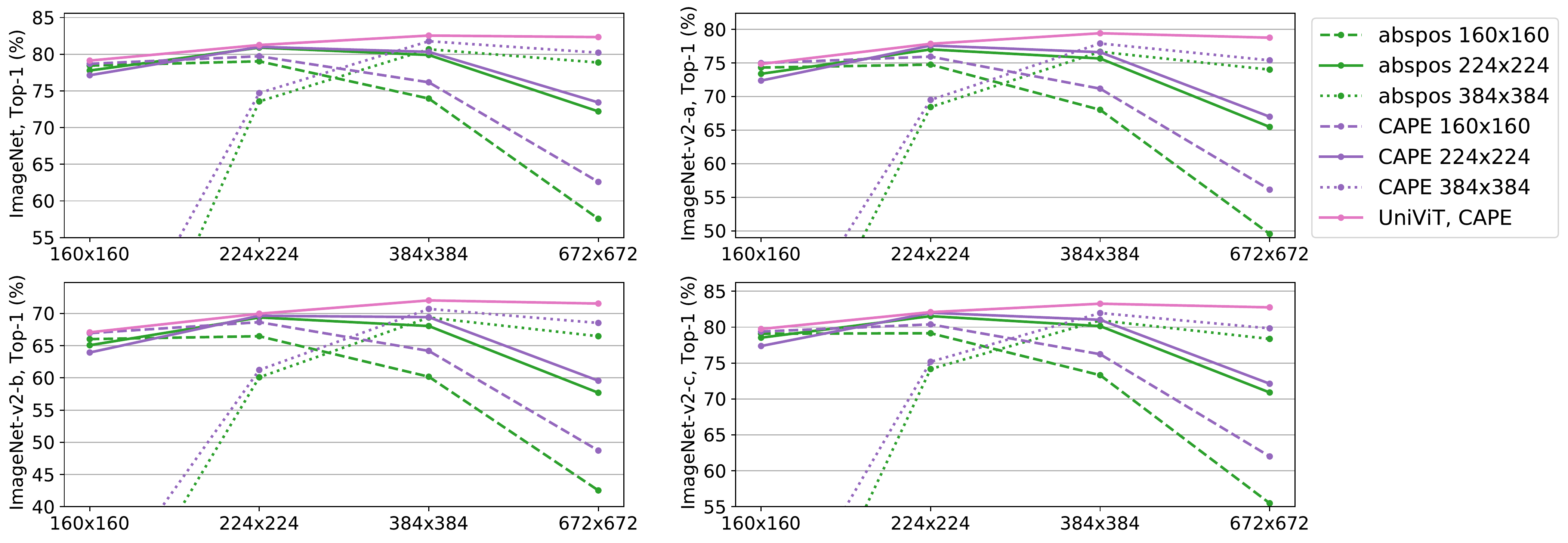}
  \caption{Top-1 accuracy on ImageNet and ImageNet-v2 for ViT models with either \textit{abspos} or CAPE trained on each particular resolution and UniViT model trained on the mixture of resolutions.\label{fig:vision:mix_sizes}}
\end{figure}

\begin{figure}
\CenterFloatBoxes
\begin{floatrow}
\ffigbox{%
  \includegraphics[width=0.4\textwidth]{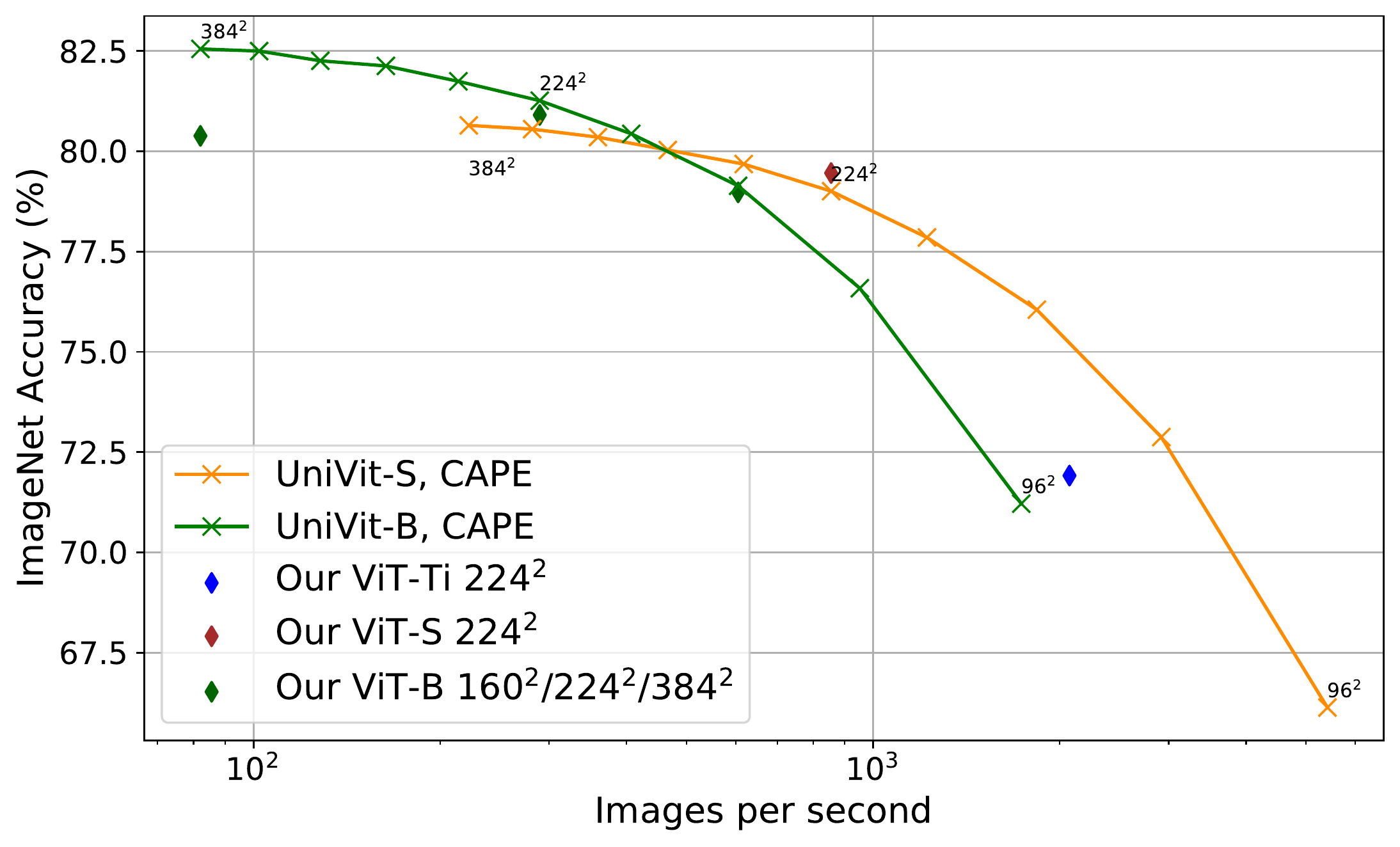}%
}{%
  \caption{Accuracy with respect to throughput on ImageNet at inference time under variation of image resolution. ``-S'' and ``-Ti'' refer to small and tiny architectures~\cite{dosovitskiy2020image}, respectively.
  \label{fig:vision:speed}}%
}
\killfloatstyle
\ttabbox
{\caption{
  Test time augmentation (TTA) results on ImageNet when predictions on resolutions $r^2$, $(r+32)^2$ and $(r-32)^2$ are combined. \textit{Sinpos} and CAPE are trained on $224^2$ resolution.\label{tab:vision_tta}}
}{
  \resizebox{0.9\linewidth}{!}{
\begin{tabular}{@{}cccccc@{}}
\toprule
\multirow{2}{*}{Model} & \multirow{2}{*}{$r$} & \multicolumn{2}{c}{Top-1 (\%)} & \multicolumn{2}{c}{Top-5 (\%)} \\
\cmidrule(lr){3-4} \cmidrule(lr){5-6}
& & -TTA & +TTA & -TTA & + TTA \\
\midrule
sinpos & 224 & 81.32 & 81.47 & 95.44 & 95.54 \\
CAPE & 224 & 81.01 & 81.34 & 95.18 & 95.49  \\
\midrule
\midrule
UniViT, sinpos & 224 & 80.82 & 81.34 & 95.40 & 95.57  \\
UniViT, CAPE & 224 & 81.26 & 81.64 & 95.56 & 95.71 \\
\midrule 
UniViT, sinpos & 384 & 82.31 & 82.44 & 96.04 & 96.14 \\
UniViT, CAPE & 384 & 82.55 & 82.72 & 96.18 & 96.22 \\
\bottomrule
\end{tabular}
}
}
\end{floatrow}
\end{figure}

As CAPE-based models can handle any image resolution, we propose a single universal ViT model, called UniViT, which is trained on images of different resolutions: during training we resize all images in a batch to a randomly chosen size,
uniformly sampled in the range $[128,320]^2$ with a step of $32$. 
For experiments with UniViT training configuration remains the same as for ViT. Because of training on the images of different resolution UniViT training time is only 1.1x longer than ViT trained on $224^2$ resolution.  
In Figure~\ref{fig:vision:mix_sizes} we compare UniViT model trained with CAPE ($\lambda=1$) against different ViT models trained on each particular resolution: UniViT outperforms single-resolution ViT models for any given resolution and, moreover, generalizes to non-training resolutions well.

Image resolution directly impacts throughput: computational complexity of attention is $O(N^4)$ for a $N\times N$ image. 
In Figure~\ref{fig:vision:speed}, we show UniViT with CAPE throughput and accuracy with respect to input image resolution.  On $96^2$ resolution UniViT with CAPE handles throughput and accuracy similar to ``tiny'' vanilla ViT, while ``small'' UniViT with CAPE has 4\% higher accuracy (with identical throughput) on resolution $160^2$ and 1.4x higher throughput (with identical accuracy) on resolution $128^2$.
Thus, UniViT unlocks the possibility of dynamically adjusting throughput of a model in a production regime under heavy loads, a practical alternative to improving model throughput at inference time via decreasing model size.

We further improve UniViT with CAPE accuracy by resizing each image to its optimal resolution at inference, as shown in Appendix~\ref{app:vision:resize} Table~\ref{tab:vision:own_size_eval}. 
We split ImageNet validation images into 8 bins, according to their size. By selecting an optimal resizing strategy in each bin we are able to improve top-1 accuracy to 82.92\% (in comparison, the model has 81.26\% on $224^2$ and 82.55\% on $384^2$).

\subsubsection{Resizing as Test Time Augmentation (TTA)}

As both \textit{sinpos}- and CAPE-based models handle well different image resolutions, we propose to perform test time resolution augmentation when evaluating a single model. 
For TTA we average model's logits evaluated on three resolutions for the same image: $r^2$, $(r-32)^2$ and $(r+32)^2$, where $r$ is either 224 or 384.
As shown in Table~\ref{tab:vision_tta}, ViT and UniViT models trained with either \textit{sinpos} or CAPE embeddings get 0.2\%-0.5\% top-1 accuracy boost with this test time augmentation.

\subsection{Automatic Speech Recognition (ASR)}\label{sec:asr}

Recently it was shown that Transformer~\cite{vaswani2017attention} architectures are state-of-the-art on different public benchmarks for ASR~\cite{zeyer2019comparison,mohamed2019transformers,likhomanenko2020rethinking,chan2021speechstew}.

\paragraph{Data} 
We evaluate our models on several English speech datasets, both on in-domain data and out-of-domain data. We also analyze our models generalization to long sequences. 
We consider two standard training benchmarks: Wall Street Journal (WSJ)~\cite{garofolo1993csr,linguistic1994csr,woodland1994large}, read speech with 81.5h of training data, and TED-LIUM v3 (TL)~\cite{hernandez2018ted}, oratory speech with 452h of training data. Besides these datasets we use two other sets for evaluation only: i) LibriSpeech (LS)~\cite{panayotov2015librispeech}, read speech from audiobook recordings (we use only test sets with clean, \emph{test-clean}, and noisy, \emph{test-other}, speech); ii) Robust Video (RV), our in-house English video dataset, which is sampled from public social media videos and aggregated and de-identified before transcription; these videos contain a diverse range of speakers, accents, topics, and acoustic conditions making ASR difficult; the test sets are composed of \emph{clean} and \emph{noisy} subsets. 
Details on data and its statistics can be found in Appendix~\ref{app:asr:data}.

\paragraph{Evaluation} 
To evaluate our acoustic models on sequence lengths not seen at training time, we split all evaluation utterances by their duration $T$ into the following bins: $T<10$s, $T\in[10-15)$s, $T\in[15, 20)$s and $T>=20$s. Our performance metric is word error rate (WER) (no language model is involved), reported for each sequence length bin and for the entire evaluation dataset. For RV data a hand-crafted segmentation is available, allowing us to evaluate on the exact same data, but segmented in different ways. More precisely, for RV data we prepare 8 sets where audios have the following respective durations: $T=10$, $15$, $20$, $25$, $30$, $35$, $40$, $45$s.

\paragraph{Acoustic Model (AM) Training}
All models are trained with Connectionist Temporal Classification~\cite{graves2006connectionist}. SpecAugment~\cite{park2019specaug} is used as data augmentation in training, and the network architecture follows~\cite{likhomanenko2020rethinking}: the AM encoder is composed of a 1D convolution (kernel 7, stride 3) with a GLU activation and 36 4-heads Transformer layers~\cite{vaswani2017attention}, finally followed by a linear layer which outputs a score for each target token.
Our token set consists of 26 English alphabet letters, augmented with the apostrophe and a word boundary token (further details in Appendix~\ref{app:asr:am}).

\paragraph{Positional Embedding}
As in vision experiments, we evaluate \textit{nopos}, \textit{sinpos}, \textit{abspos} and CAPE-based models. In addition, we evaluate models trained with relative positional embeddings: in that case, no absolute positions are used, and learnable relative positional embeddings~\cite{shaw2018self} (\textit{relpos}) are trained in each Transformer layer. 
We follow~\cite{likhomanenko2020rethinking} to train an AM baseline with~\textit{relpos}. For models with other positional embeddings, the training \textit{configuration remains identical}. \textit{Abspos} $\{\mathbf{E}(t)\}_{t=1}^N$ is set to cover 13.8s of context. 
At training/evaluation time for the longer sequences we define \textit{abspos} as $\mathbf{E}(t) \equiv  \mathbf{E}{(t \mod N)}$ for $t > N$. 
This extrapolation at training time leaves a chance to the acoustic model to generalize to unseen (longer) sequence lengths. 
\textit{Relpos} spans a large context, 26.8s to the left/right. 
CAPE's global shift covers 60s, while a local shift is set to its maximum to preserve the frames order; $\lambda_{max}=1.1$ and $\lambda_{max}=2$ for WSJ and TL, respectively.

\subsubsection{Results}

\begin{figure}[t!]
  \centering
  \includegraphics[width=0.86\textwidth]{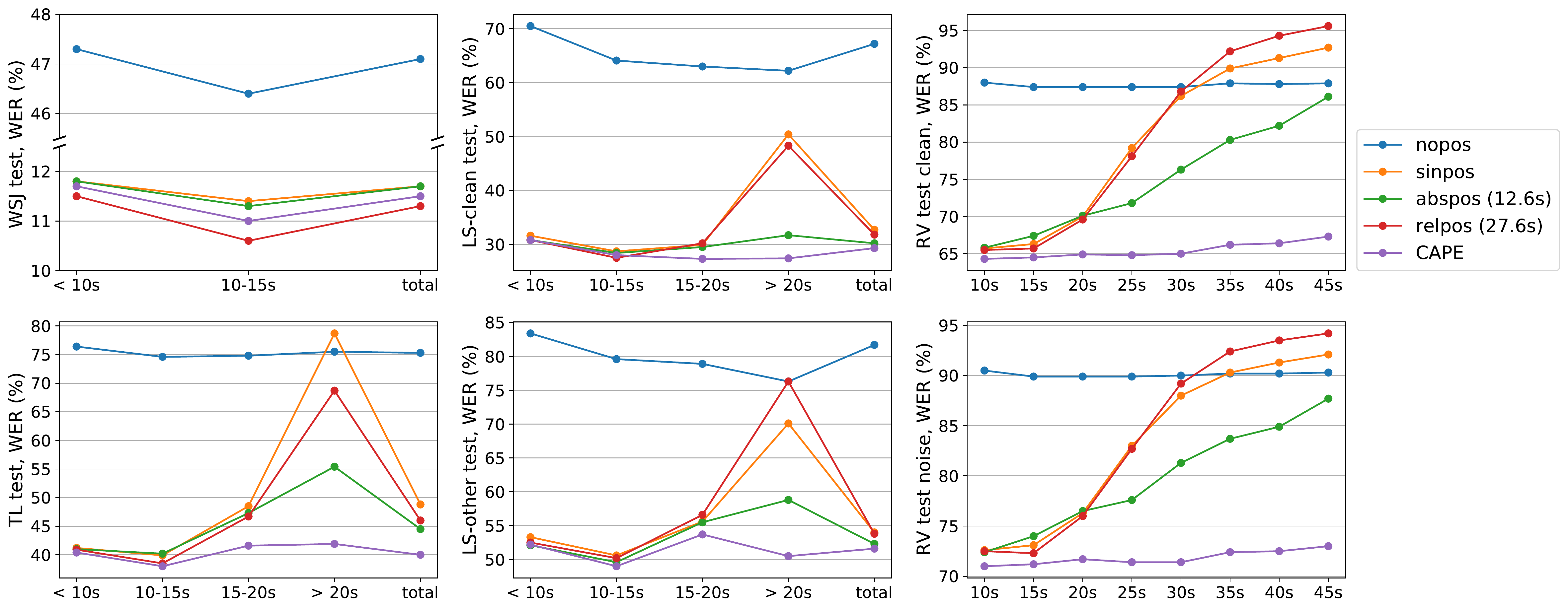}
  \caption{Word error rate for models trained on WSJ with different positional embeddings.\label{fig:speech:wsj}}
\end{figure}

\begin{figure}[t!]
  \centering
  \includegraphics[width=0.86\textwidth]{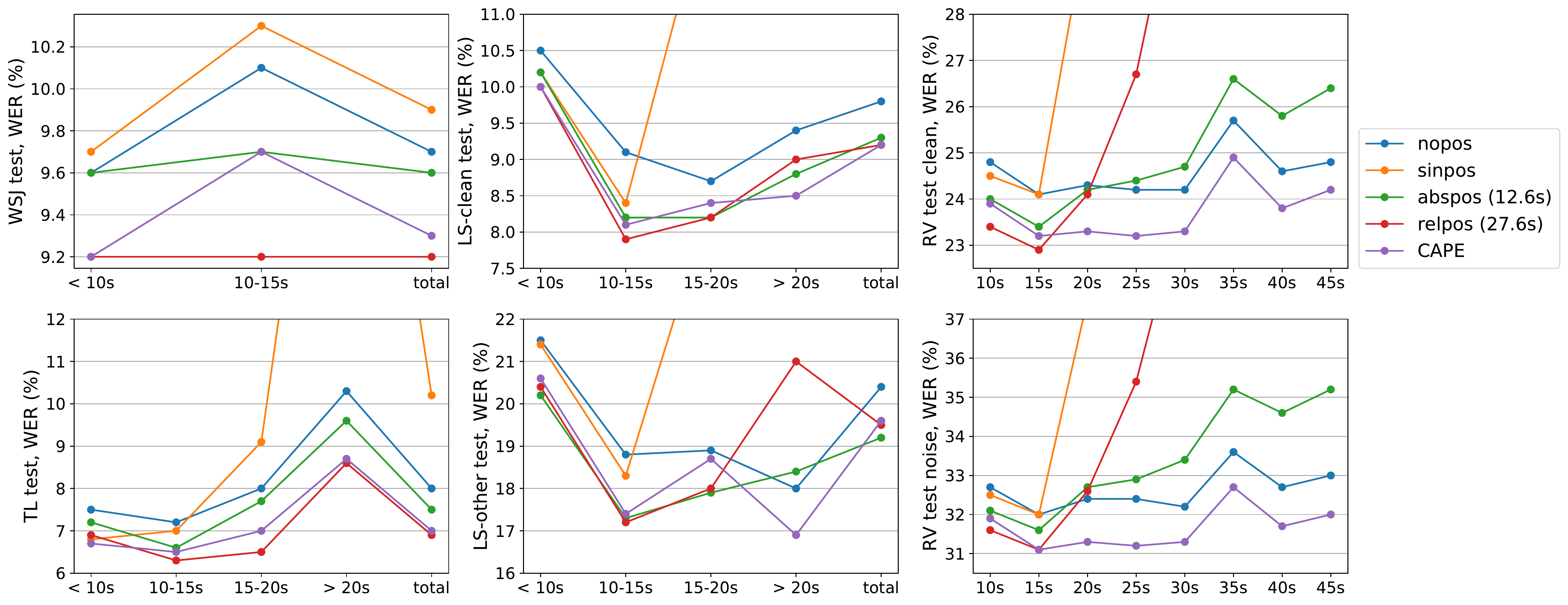}
  \caption{Word error rate for models trained on TED-LIUM v3 with different positional embeddings.\label{fig:speech:tl}}
\end{figure}

A model trained on WSJ with CAPE outperforms other positional embeddings on both public and RV data across different audio durations, as shown in Figure~\ref{fig:speech:wsj}. A model trained on TL with CAPE outperforms \textit{nopos} and \textit{sinpos} on all data, outperforms \textit{abspos} and \textit{relpos} for audio longer than 20s, and behaves similarly on shorter durations, see Figure~\ref{fig:speech:tl}. On RV data, CAPE-based models perform uniformly well on different audio durations, including long ones. In contrast, other embeddings-based models are seriously impacted when audio duration increases.
Finally, CAPE does not have computational or parameters overheads compared to \textit{relpos}.

\paragraph{No Positional Embedding} 
As expected, \textit{nopos} models (both WSJ and TL ones) perform similarly in WER across different audio durations. However, \textit{nopos} TL model performs surprisingly well: it is competitive to positional embeddings-based models on public data. On RV data, \textit{nopos} TL model outperforms all other models, except CAPE when $T>20$s. 
We perform ablations in Appendix~\ref{app:asr:ablation} to show that key ingredients are CTC loss, sufficient model capacity, and large amount of data for this effect to occur.

\begin{figure}
\CenterFloatBoxes
\begin{floatrow}
\ffigbox{%
  \includegraphics[width=0.5\textwidth]{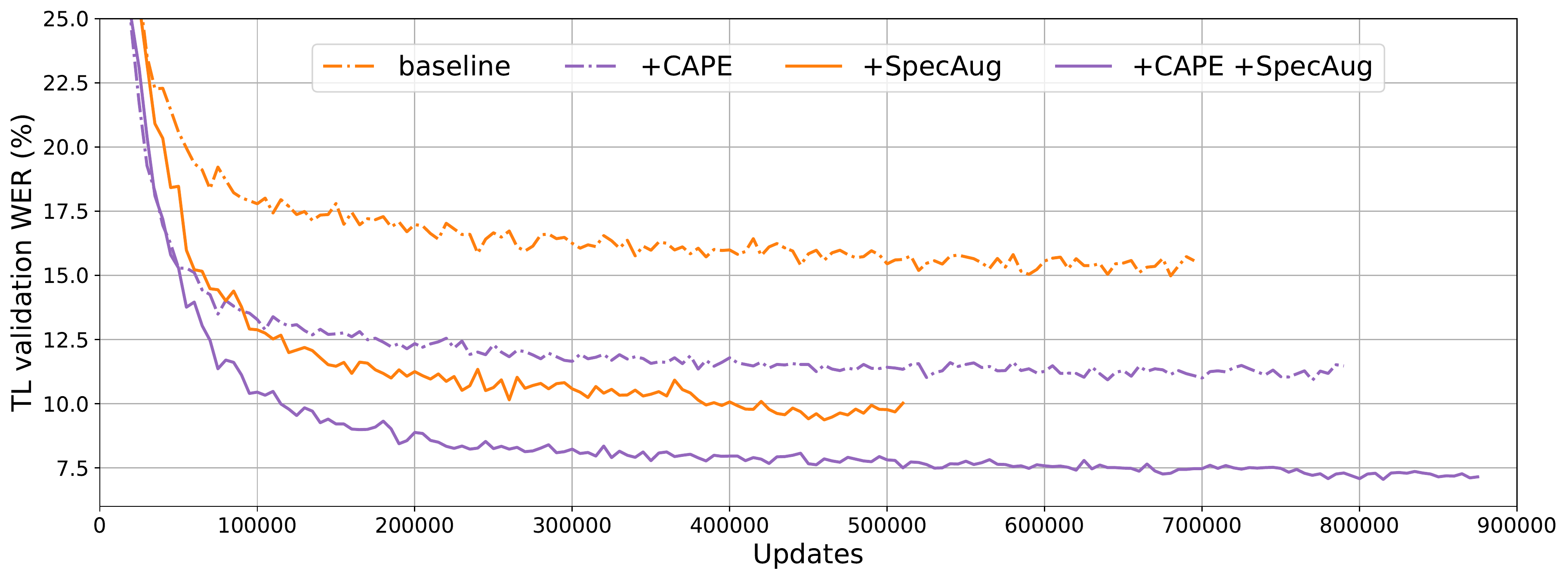}
  \caption{Validation WER for models trained with different augmentations: ``baseline'' is a model with \textit{sinpos}, ``+CAPE'' adds CAPE's global and local shifts, ``+SpecAug'' adds SpecAugment.\label{fig:asr_aug}}
 }
\killfloatstyle
\ttabbox{
    \caption{WMT’14 BLEU score (3 runs avg.).\label{tab:mt}}
}{
  \resizebox{0.95\linewidth}{!}{
\begin{tabular}{@{}cccc@{}}
\toprule
Model \& Lang. & Embedding & Valid BLEU & Test BLEU \\
\midrule
\multirow{3}{*}{6L-6L} & sinpos & 26.88$\pm$0.05  & 27.66$\pm$0.10 \\
 & abspos & 26.68$\pm$0.05 & 27.36$\pm$0.06 \\
 & relpos & 26.81$\pm$0.16 & 27.92$\pm$0.07 \\
\multirow{2}{*}{DE} & CAPE, $\Delta = 5$ & 26.86$\pm$0.13 & 27.89$\pm$0.07 \\
 & CAPE, $\Delta = 50$ & 27.09$\pm$0.03 & 27.77$\pm$0.16
 \\
\midrule
\midrule
\multirow{2}{*}{18L-18L} & sinpos & 27.09$\pm$0.06 & 28.28$\pm$0.28 \\
 & abspos & 27.23$\pm$0.02 & 28.26$\pm$0.22 \\
 DE & CAPE, $\Delta = 10$ & 27.17$\pm$0.10 & 28.44$\pm$0.06 \\
\midrule
\midrule
\multirow{3}{*}{6L-6L} & sinpos & 47.27$\pm$0.03 & 41.13$\pm$0.07 \\
& abspos & 47.22$\pm$0.03 & 41.21$\pm$0.04 \\
& relpos & 47.12$\pm$0.03 & 41.33$\pm$0.13 \\
\multirow{2}{*}{FR} & CAPE, $\Delta = 5$ & 47.22$\pm$0.03 & 41.59$\pm$0.03 \\
& CAPE, $\Delta = 50$ & 47.14$\pm$0.02 & 41.48$\pm$0.10 \\
\bottomrule
\end{tabular}
}
}
\end{floatrow}
\end{figure}

\paragraph{CAPE as Augmentation}
CAPE can be viewed as a data augmentation performed on input data (positions), which regularizes the trained model. 
We demonstrate this by training on TL with either \textit{sinpos} or CAPE and with/without SpecAugment (no other augmentations are used), see Figure~\ref{fig:asr_aug}. Baseline \textit{sinpos} without any augmentation performs the worst with a large gap. Including either CAPE or SpecAugment decreases the WER significantly by 4-5\%: SpecAugment is more beneficial due to its domain relevance.
Combining together CAPE and SpecAugment further decreases the WER by 2.5-3.5\%,
showing that augmentations are complementary to each other.

\subsubsection{Padding-free ASR with CAPE and Variable STFT Hop Distance}

In ASR, when batching audios of different duration, one often relies on padding tokens. We propose instead to perform time stretching augmentation on all audios in the batch, such that they will have the same number of frames. We perform this augmentation by tuning the short-time Fourier Transform (STFT) hop distance when computing the audio features. Positions remain tied to the original audio timestamps. 
These models trained either on WSJ or TL show better WER across the board, compared to models trained with a padding approach, as shown in Appendix~\ref{app:asr:hop} Figures~\ref{fig:abl:wsj_hop} and~\ref{fig:abl:tl_hop}, and improvements on RV are quite consistent. As before, CAPE-based models outperform \textit{sinpos} models.

We found this padding-free approach convenient, as it \textit{alleviates the implementation of special cases to handle padding tokens} in ASR models: e.g. any normalization should be aware which tokens are padding, otherwise normalization constants would depend on the amount of padding; any attention module should be aware which tokens it should not attend to.
Moreover, for efficient computations and reducing padding tokens audio samples are often packed together via sorting by their duration; this reduces variability in the batches drastically. 
Our results demonstrate that with CAPE and padding-free approach we can mix samples of not-too-different lengths within a batch, providing better randomization and utilize all frames. 
While UniViT adjusts number of tokens by changing resolution, the STFT hop distance achieves the same for audio.
By adjusting the hop distance during inference for padding-free ASR, we can achieve higher throughput with similar recognition quality, see Appendix~\ref{app:asr:hop} Figure~\ref{fig:abl:tl_hop_time}. However, very low resolution (e.g. 128x128) in images works well enough while speech recognition is far more sensitive to token sparsification as phoneme can last as little as 30-50ms.

\subsection{Machine Translation (MT)}\label{sec:mt}
Our MT experiments follow the recent results with Transformers combined with a new initialization scheme (ADMIN)~\cite{liu2019deep,liu2020very}. This approach allows to train very deep state-of-the-art Transformers for MT. 
We did not implement back-translation or other domain-specific augmentations.

\paragraph{Data and Models Training} 
Experiments are conducted on standard WMT’14 English-French (FR)  and  English-German (DE) benchmarks. For both benchmarks we follow~\cite{liu2019deep,liu2020very}: for FR we use a 40k subword vocabulary, and evaluate on the provided ``valid'' file for validation and newstest14 for test. On DE, we consider a 32K subword vocabulary, newstest2013 for validation, and newstest2014 for test. We reproduce results from~\cite{liu2019deep,liu2020very} by training a~\textit{sinpos}-based model with 6L-6L, 18L-18L for DE and 6L-6L for FR encoder-decoder layers with ADMIN. \textit{Training configuration stays the same} for other positional embedding-based models, other than positional embeddings being either \textit{abspos} (covers 1k tokens), \textit{relpos} (learnable \cite{shaw2018self}, covers max train content: 150 left/right tokens) or CAPE in both encoder and decoder layers. 
For CAPE to have approximate correspondence in positions of source and target sentences, we first scale positions of source language by a factor $\alpha = \frac{\text{\# tokens in target corpus}}{\text{\# tokens in source corpus}} \in\mathbb{R}$ so that positions interval is loosely matched between source and target sentences. For each training sample we then apply the same global shift and scale for source and target positions. Local shifts for source/target positions are independently sampled from $\mathcal{U}(-0.5, 0.5)$. As no absolute positions are provided anymore, we prepend source sentences with a ``begin of sentence" token to hint the decoder with a first position during both training and evaluation.

\paragraph{Evaluation}
We select the best checkpoint according to BLEU on the validation set, using a beam size 4 for DE and 5 for FR. 
Following convention, BLEU is computed by \url{multi-bleu.perl} via the standardized tokenization of the publicly-accessible dataset.
As WMT’14 test sets have limited length of sentences being same as train we artificially stack together sentences in test sets (source and target) to have 300+ tokens in each new target sample (3k original pairs of sentences are transformed into 270 pairs): during stacking dots on junctions are replaced with commas and first letters of the next stacked sentence are lower-cased to match one-sentence setup. On these test sets we compute log loss for each position: lower value shows that a model is more likely to predict a correct word at a particular position.

\paragraph{Results}

\begin{figure}[t!]
  \centering
  \includegraphics[width=0.9\textwidth]{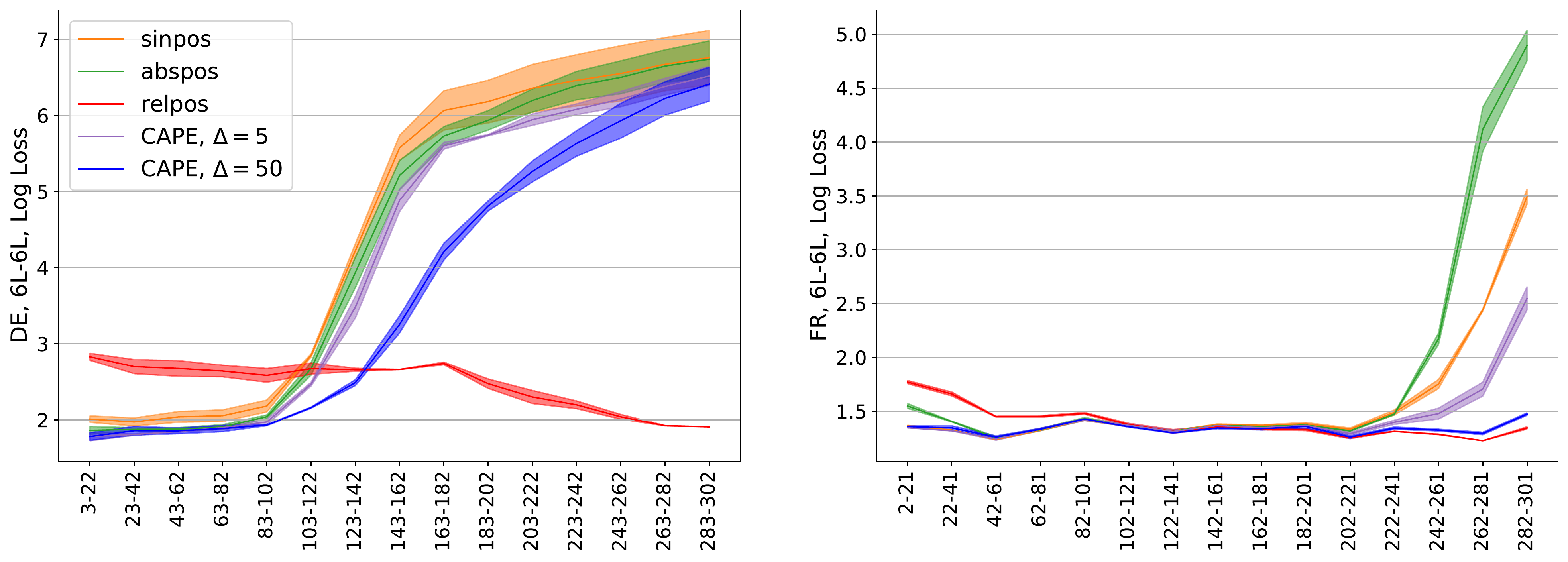}
  \caption{Average negative log loss across consecutive positions measured on test sets with stacked sentences for models with different positional embeddings: 6L-6L DE (left) and 6L-6L FR (right). \label{fig:mt}}
\end{figure}

Comparison between models trained with different positional embeddings on WMT'14 benchmarks is shown in Table~\ref{tab:mt}. CAPE outperforms \textit{sinpos} and \textit{abspos} on all settings, is similar to \textit{relpos} on DE and outperforms it on FR.
Figure~\ref{fig:mt} shows that i) for positions covered in training (<150) all absolute positional embeddings behave similarly and outperform \textit{relpos}; ii) for positions not seen during training (>200) \textit{relpos} outperforms others. However, CAPE generalizes better than \textit{abspos} and \textit{sinpos} and, moreover, is able to generalize similar to \textit{relpos} with more data (FR).
 
\section{Discussion and Conclusion}\label{sec:discussion}

Encoding positional information is a key component of attention-based models.
Poor generalization of absolute sinusoidal positional embeddings led to numerous works investigating ways to encode relevant positional information. Existing solutions are often modality-specific, non-trivial to implement and incur computational overhead. 
We demonstrated in different domains that existing positional embeddings may generalize poorly in certain conditions. We introduced a simple and efficient continuous augmented positional embedding, CAPE, which
preserves some information about relative token positions.
Thanks to its continuous nature, CAPE allows augmentations which were previously not possible.
CAPE makes models more flexible both at training and inference. It generalizes well to input sizes not seen during training across a variety of domains.
We expect emergence of new training and production pipelines 
that leverage the adjustable throughput property when tuning the input size: CAPE provides a relatively simple way of producing more efficient Transformer by down-sampling input (e.g for image and audio).
Going further, CAPE-based architectures are free from baked-in restrictions on patches positions: these could overlap, or be sparse for example -- opportunities impossible for convolution-containing Transformers.
In contrast to relative positional embeddings which modify attention mechanism, CAPE is ready to be used by novel attention mechanisms, such as ``linear''  Transformers~\cite{tay2021long}.
Finally, CAPE can be combined with relative positional embeddings like~\cite{dai2019transformer} and~\cite{su2021roformer} to limit over-fitting to exact relative positions.

\paragraph{Limitations}
CAPE applies only to attention-based models, and no testing was performed outside described modalities. 
From a representation perspective we demonstrated that CAPE is capable of providing relative positioning, however, a model should ``learn'' it. 
Thus, we can expect that relative positional embeddings should be beneficial in settings with small amount of data because of more appropriate inductive bias. 
Proposed UniViT model was tested only for image recognition task and further exploration of broader UniViT applicability to other tasks is a subject of future work. 

\section{Acknowledgments}

We would like to thank Mark Tygert and Edouard Grave for relevant references and helpful discussions, Paden Tomasello for English language editing.

\bibliographystyle{plain}
\bibliography{mybib}

\newpage
\appendix

\section{CAPE Implementation in Python}\label{app:impl}

\begin{lstlisting}
import numpy as np

def augment_positions_1d(
    positions_1d: np.ndarray, 
    mean_normalize: bool,
    augment: bool,    # True during training
    max_global_shift, # delta max  
    max_local_shift,  # epsilon max  
    max_scale,        # lambda max
    rng=np.random.RandomState(42)
):
    """
    Takes original positions, returns modified ones.
    Can reuse sin/cos embedding from "Attention is all you need".
    Code handles NaNs is positions_1d input as if those correspond to pad tokens
    """
    assert max_scale >= 1
    batch_size, n_tokens = positions_1d.shape
    if mean_normalize:
        positions_1d -= np.nanmean(positions_1d, axis=1, keepdims=True)
    if augment:
        delta = rng.uniform(-max_global_shift, +max_global_shift, size=[batch_size, 1])
        delta_local = rng.uniform(-max_local_shift, +max_local_shift, size=[batch_size, n_tokens])
        log_lambdas = rng.uniform(-np.log(max_scale), +np.log(max_scale), size=[batch_size, 1])
        new_positions = (positions_1d + delta + delta_local) * np.exp(log_lambdas)
        return new_positions
    else:
        return positions_1d

def CAPE_2d(
    n_patches: int,   # number of patches, default in ViT is 14 
    batch_size: int,
    augment: bool,    # True during training
    n_channels: int,  # embedding size for one patch
    max_global_shift, # delta max  
    max_local_shift,  # epsilon max  
    max_scale,        # lambda max
    rng=np.random.RandomState(42)
):
    """Prepares grid of CAPE embeddings for provided grid size"""
    x = np.zeros([batch_size, n_patches, n_patches])
    y = np.zeros([batch_size, n_patches, n_patches])
    x += np.linspace(-1, 1, n_patches)[None, :, None]
    y += np.linspace(-1, 1, n_patches)[None, None, :]
    if augment:
        # global shift
        x += rng.uniform(-max_global_shift, +max_global_shift, size=[batch_size, 1, 1]) 
        y += rng.uniform(-max_global_shift, +max_global_shift, size=[batch_size, 1, 1]) 
        # local shift
        x += rng.uniform(-max_local_shift, +max_local_shift, size=x.shape)
        y += rng.uniform(-max_local_shift, +max_local_shift, size=y.shape)
        # scaling
        lambdas = np.exp(rng.uniform(-np.log(max_scale), + np.log(max_scale), size=[batch_size, 1, 1]))
        x *= lambdas
        y *= lambdas

    assert n_channels % 2 == 0
    half_channels = n_channels // 2
    rho = 10 ** (np.arange(1, half_channels + 1) / half_channels)
    # recommended simpler approximate implementation 
    # rho = 10 ** np.linspace(0, 1, half_channels)
    w_x = rho * np.cos(np.arange(half_channels))
    w_y = rho * np.sin(np.arange(half_channels))
    
    phase = np.pi * (w_x * x[:, :, :, None] + w_y * y[:, :, :, None])
    return np.concatenate([np.cos(phase), np.sin(phase)], axis=-1)
\end{lstlisting}

\newpage 
\section{Image Recognition Experiments}
\subsection{Technical Details}\label{app:vision:tech}

\begin{table}[h!]
\caption{Top-1 accuracy (\%) for ViT and UniViT models evaluated on ImageNet validation set and ImageNet-v2 test sets with images resized to different resolutions: $160^2$, $224^2$, $384^2$ and $672^2$. Models trained on $224^2$ and further fine-tuned on $384^2$ resolution are marked with ``+ft''. ``-S'' and ``-Ti'' refer to small and tiny architectures~\cite{dosovitskiy2020image}, respectively.\label{tab:vision_all_top1}}
\begin{center}
\npdecimalsign{.}
\nprounddigits{2}
\npnoroundexp
\resizebox{\linewidth}{!}{
\begin{tabular}{@{}lcn{2}{2}n{2}{2}n{2}{2}n{2}{2}n{2}{2}n{2}{2}n{2}{2}n{2}{2}n{2}{2}n{2}{2}n{2}{2}n{2}{2}n{2}{2}n{2}{2}n{2}{2}n{2}{2}@{}}
\toprule
\multirow{2}{*}{Model} & Train & \multicolumn{4}{c}{ImageNet} & \multicolumn{4}{c}{ImageNet-v2-a} & \multicolumn{4}{c}{ImageNet-v2-b} & \multicolumn{4}{c}{ImageNet-v2-c} \\
\cmidrule(lr){3-6} \cmidrule(lr){7-10} \cmidrule(lr){11-14} \cmidrule(lr){15-18}
& Res. & \texttt{$160^2$} & \texttt{$224^2$} & \texttt{$384^2$} & \texttt{$672^2$} & \texttt{$160^2$} & \texttt{$224^2$} & \texttt{$384^2$} & \texttt{$672^2$} & \texttt{$160^2$} & \texttt{$224^2$} & \texttt{$384^2$} & \texttt{$672^2$} & \texttt{$160^2$} & \texttt{$224^2$} & \texttt{$384^2$} & \texttt{$672^2$} \\
\midrule
nopos &  \multirow{5}{*}{$160^2$} & 77.334 & 78.38 & 75.462 & 61.004 & 72.39 & 73.78 & 69.65 & 52.99 & 64.41 & 66.25 & 62.58 & 46.58 & 77.67 & 78.76 & 75.28 & 59.13 \\
abspos &  & 78.454 & 79.038 & 73.96 & 57.562 & 74.31 & 74.75 & 68.02 & 49.54 & 66.02 & 66.47 & 60.18 & 42.53 & 79.08 & 79.15 & 73.31 & 55.48 \\
sinpos &  & 79.052 & 74.382 & 65.654 & 44.128 & 74.75 & 70.11 & 59.61 & 36.75 & 66.91 & 62.13 & 52.28 & 30.55 & 79.37 & 75.41 & 65.45 & 42.51 \\
CAPE, $\lambda=1$ &  & 78.738 & 79.694 & 75.82 & 61.868 & 74.94 & 75.75 & 70.5 & 54.91 & 66.56 & 68.17 & 62.78 & 46.74 & 79.51 & 80.18 & 75.71 & 60.39 \\
CAPE &  & 78.702 & 79.73 & 76.176 & 62.59 & 74.98 & 75.94 & 71.17 & 56.13 & 66.95 & 68.65 & 64.19 & 48.72 & 79.38 & 80.38 & 76.23 & 62 \\
\midrule
nopos & \multirow{9}{*}{$224^2$} & 75.296 & 78.966 & 78.678 & 72.922 & 70.52 & 74.79 & 74.15 & 65.99 & 61.59 & 66.93 & 67.01 & 59.41 & 75.94 & 79.5 & 78.81 & 71.06 \\
abspos &  & 77.782 & 80.904 & 79.898 & 72.206 & 73.38 & 77.01 & 75.66 & 65.47 & 65.07 & 69.38 & 68.05 & 57.69 & 78.53 & 81.53 & 80.14 & 70.9  \\
sinpos &  & 77.152 & 81.316 & 79.722 & 70.714 & 72.91 & 77.52 & 75.48 & 63.74 & 64.45 & 70.14 & 67.96 & 55.67 & 78.3 & 82.19 & 80.13 & 69.28 \\
CAPE, $\lambda=1, \Delta=0$ & & 77.532 & 81.084 & 80.18 & 72.344 & 73.41 & 77.61 & 75.81 & 65.99 & 64.85 & 70.23 & 68.22 & 57.83 & 78.33 & 82.05 & 80.06 & 70.99 \\
CAPE, $\lambda=1, \epsilon=0$ & & 77.458 & 81.144 & 80.488 & 72.132 & 73.35 & 77.9 & 76.36 & 64.95 & 64.84 & 69.84 & 68.45 & 57.31 & 78.8 & 81.84 & 80.58 & 70.53 \\
CAPE, $\lambda=1$ &  & 77.698 & 81.01 & 80.384 & 73.064 & 72.88 & 77.67 & 76.23 & 67.35 & 64.79 & 70.25 & 69.36 & 59.59 & 78.16 & 82.22 & 80.96 & 72.07 \\
CAPE, $\Delta=0$ & & 77.35 & 81.08 & 80.502 & 73.108 & 73.05 & 77.59 & 76.73 & 67.25 & 64.98 & 69.75 & 69.14 & 59.01 & 78.32 & 81.88 & 81 & 72.04 \\
CAPE, $\epsilon=0$ & & 77.706 & 81.296 & 80.572 & 73.354 & 73.51 & 77.71 & 76.72 & 66.58 & 65.01 & 69.93 & 69.59 & 59.31 & 78.02 & 81.93 & 80.86 & 71.73 \\
CAPE &  & 77.138 & 81.006 & 80.326 & 73.426 & 72.37 & 77.59 & 76.6 & 66.99 & 63.94 & 69.65 & 69.43 & 59.56 & 77.37 & 82 & 81.02 & 72.12 \\
\midrule
abspos & \multirow{4}{*}{$384^2$} & 21.826 & 73.57 & 80.682 & 78.87 & 19.99 & 68.43 & 76.66 & 74 & 16.23 & 60.09 & 69.37 & 66.47 & 24.02 & 74.17 & 80.94 & 78.36 \\
sinpos &  & 7.708 & 75.55 & 82.416 & 80.734 & 7.16 & 70.87 & 78.82 & 76.21 & 5.26 & 62.29 & 71.32 & 68.82 & 8.72 & 75.57 & 82.82 & 80.47 \\
CAPE, $\lambda=1$ &  & 31.978 & 75.378 & 82.554 & 80.94 & 28.47 & 71 & 78.97 & 75.85 & 23.51 & 62.4 & 71.55 & 69.02 & 34.15 & 76.25 & 82.77 & 80.6 \\
CAPE &  & 32.97 & 74.712 & 81.778 & 80.216 & 29.23 & 69.5 & 77.9 & 75.39 & 24.21 & 61.23 & 70.71 & 68.53 & 34.83 & 75.18 & 81.96 & 79.83 \\
\midrule
nopos+ft & \multirow{5}{*}{ \shortstack{ $224^2$ \\ $\downarrow$ \\ $384^2$} } & 46.104 & 75.74 & 80.376 & 78.894 & 41.42 & 71.72 & 76.69 & 74.16 & 33.82 & 62.81 & 69.04 & 67.06 & 47.87 & 76.58 & 80.92 & 78.69 \\
abspos+ft &  & 34.164 & 78.51 & 82.346 & 80.632 & 29.71 & 75.02 & 79.1 & 76.1 & 24.45 & 66.54 & 71.91 & 68.79 & 35.76 & 79.33 & 83.15 & 80.63 \\
sinpos+ft &  & 25.12 & 77.694 & 82.772 & 81.016 & 22.61 & 73.35 & 79.42 & 76.5 & 17.77 & 65.27 & 72.16 & 69.1 & 26.62 & 78.72 & 83.33 & 80.68 \\
CAPE+ft, $\lambda=1$ &  & 57.8 & 78.628 & 82.668 & 81.284 & 53.25 & 74.87 & 79.19 & 77.3 & 44.91 & 67.07 & 72.21 & 70.2 & 59.69 & 79.72 & 83.6 & 81.59 \\
CAPE+ft &  & 58.462 & 78.136 & 82.462 & 80.936 & 53.55 & 73.89 & 79.4 & 76.89 & 44.82 & 65.3 & 71.74 & 69.71 & 60.05 & 78.76 & 83.08 & 81.53 \\
\midrule
UniViT, sinpos & \multirow{3}{*}{mix} & 78.944  & 80.824  & 82.306  & 82.118  & 74.64  & 77.21  & 78.58  & 78.29  & 66.57  & 69.74  & 71.48  & 71.42  & 79.53  & 81.67  & 82.82  & 82.53  \\
UniViT, CAPE $\lambda=1$ &  & 79.136  & 81.26  & 82.55  & 82.34  & 74.85  & 77.85  & 79.42  & 78.76  & 67.07  & 69.97  & 72.03  & 71.54  & 79.74  & 82.09  & 83.26  & 82.75  \\
UniViT, CAPE &  & 79.05  & 81.164  & 82.282  & 81.834  & 74.88  & 77.5  & 78.96  & 77.87  & 67.08  & 69.88  & 72.01  & 70.99  & 79.78  & 82.06  & 83.1  & 82.21  \\
\midrule
abspos-S & $224^2$ & 74.888 & 79.456 & 77.828 & 64.318 & 70.89 & 75.83 & 73.45 & 57.4 & 62.04 & 68.12 & 65.86 & 49.22 & 76.15 & 80.54 & 78.52 & 63.24 \\
UniViT-S, CAPE, $\lambda=1$ & mix & 76.05 & 79.004 & 80.644 & 80.308 & 72.57 & 75.56 & 77.59 & 76.66 & 64 & 67.31 & 70.25 & 69.41 & 78.13 & 80.44 & 81.98 & 81.3 \\
\midrule 
abspos-Ti & $224^2$ & 64.758 & 71.914 & 70.208 & 56.122 & 61.03 & 68.78 & 66.3 & 49.62 & 51.88 & 59.67 & 58.09 & 42.71 & 67.97 & 74.13 & 71.92 & 55.63 \\
UniViT-Ti, CAPE, $\lambda=1$ & mix & 65.246 & 69.826 & 72.438 & 71.152 & 62.2 & 66.4 & 68.99 & 67.45 & 53.35 & 57.25 & 61.3 & 59.72 & 69.15 & 72.64 & 74.93 & 73.23 \\
\bottomrule
\end{tabular}
}
\end{center}
\end{table}

\begin{table}[h!]
\caption{Top-5 accuracy (\%) for ViT and UniViT models evaluated on ImageNet validation set and ImageNet-v2 test sets with images resized to different resolutions: $160^2$, $224^2$, $384^2$ and $672^2$. Models trained on $224^2$ and further fine-tuned on $384^2$ resolution are marked with ``+ft''. ``-S'' and ``-Ti'' refer to small and tiny architectures~\cite{dosovitskiy2020image}, respectively.\label{tab:vision_all_top5}}
\npdecimalsign{.}
\nprounddigits{2}
\npnoroundexp
\begin{center}
\resizebox{\linewidth}{!}{
\begin{tabular}{@{}lcn{2}{2}n{2}{2}n{2}{2}n{2}{2}n{2}{2}n{2}{2}n{2}{2}n{2}{2}n{2}{2}n{2}{2}n{2}{2}n{2}{2}n{2}{2}n{2}{2}n{2}{2}n{2}{2}@{}}
\toprule
\multirow{2}{*}{Model} & Train & \multicolumn{4}{c}{ImageNet} & \multicolumn{4}{c}{ImageNet-v2-a} & \multicolumn{4}{c}{ImageNet-v2-b} & \multicolumn{4}{c}{ImageNet-v2-c} \\
\cmidrule(lr){3-6} \cmidrule(lr){7-10} \cmidrule(lr){11-14} \cmidrule(lr){15-18}
& Res. & \texttt{$160^2$} & \texttt{$224^2$} & \texttt{$384^2$} & \texttt{$672^2$} & \texttt{$160^2$} & \texttt{$224^2$} & \texttt{$384^2$} & \texttt{$672^2$} & \texttt{$160^2$} & \texttt{$224^2$} & \texttt{$384^2$} & \texttt{$672^2$} & \texttt{$160^2$} & \texttt{$224^2$} & \texttt{$384^2$} & \texttt{$672^2$} \\
\midrule
nopos & \multirow{5}{*}{$160^2$}  & 93.198 & 93.992 & 92.43 & 83.688 & 91.71 & 92.64 & 90.22 & 77.75 & 84.88 & 86.39 & 83.96 & 71.52 & 94.6 & 95.02 & 93.09 & 82.41 \\
abspos &  & 94.062 & 94.366 & 91.6 & 81.26 & 92.66 & 93.14 & 88.97 & 75.19 & 86.07 & 86.76 & 82.27 & 67.75 & 95.3 & 95.31 & 92.27 & 80.53 \\
sinpos &  & 94.264 & 91.8 & 86.414 & 68.552 & 93.2 & 90.08 & 82.75 & 61.43 & 86.65 & 83.28 & 75.62 & 54.14 & 95.7 & 93.24 & 87.03 & 67.55 \\
CAPE, $\lambda=1$ &  & 94.194 & 94.776 & 92.916 & 84.574 & 92.84 & 93.46 & 90.55 & 79.44 & 86.87 & 87.75 & 84.62 & 72.17 & 95.44 & 95.8 & 93.85 & 84.43 \\
CAPE &  & 94.148 & 94.8 & 93.026 & 85.322 & 92.95 & 93.65 & 90.82 & 80.48 & 86.68 & 87.86 & 85.22 & 73.47 & 95.27 & 95.72 & 94.06 & 84.92 \\
\midrule
nopos & \multirow{9}{*}{$224^2$}  & 92.138 & 94.222 & 94.114 & 90.89 & 90.17 & 92.79 & 92.73 & 87.58 & 82.72 & 86.92 & 87.27 & 81.61 & 93.48 & 95.19 & 94.88 & 90.91 \\
abspos &  &  93.446 & 95.26 & 94.714 & 90.488 & 91.8 & 93.91 & 93.24 & 87.13 & 85.34 & 88.52 & 87.77 & 80.23 & 94.52 & 96.17 & 95.43 & 90.76 \\
sinpos &  & 93.286 & 95.444 & 94.524 & 89.772 & 91.48 & 94.22 & 92.84 & 85.33 & 84.52 & 88.75 & 87.22 & 78.37 & 94.48 & 96.46 & 95.23 & 89.03 \\
CAPE, $\lambda=1, \Delta=0$ & & 93.294 & 95.212 & 94.73 & 90.558 & 91.47 & 93.73 & 92.91 & 87.42 & 84.8 & 88.26 & 87.69 & 80.74 & 94.43 & 96.14 & 95.37 & 90.6 \\
CAPE, $\lambda=1, \epsilon=0$ & & 93.372 & 95.416 & 94.966 & 90.722 & 91.46 & 94.29 & 93.26 & 86.82 & 85.26 & 88.87 & 88.15 & 80.26 & 94.47 & 96.22 & 95.61 & 90.34 \\
CAPE, $\lambda=1$ &  & 93.446 & 95.452 & 95.094 & 91.5 & 91.69 & 94.11 & 93.64 & 87.84 & 85.61 & 88.87 & 88.52 & 81.79 & 94.63 & 96.18 & 95.91 & 91.6 \\
CAPE, $\Delta=0$ & & 93.19 & 95.42 & 94.998 & 91.304 & 91.94 & 94.38 & 93.75 & 88.13 & 85.13 & 88.97 & 88.52 & 82.01 & 94.63 & 96.27 & 96.01 & 91.4 \\
CAPE, $\epsilon=0$ & & 93.256 & 95.318 & 94.942 & 91.246 & 91.92 & 94.24 & 93.63 & 87.8 & 84.88 & 88.64 & 88.26 & 81.1 & 94.65 & 96.18 & 95.51 & 90.99 \\
CAPE &  & 93.178 & 95.176 & 94.938 & 91.574 & 91.64 & 94.23 & 93.77 & 88.56 & 85.18 & 88.31 & 88.33 & 82.04 & 94.49 & 96.39 & 95.94 & 91.85 \\
\midrule
abspos & \multirow{4}{*}{$384^2$} & 38.446 & 90.688 & 94.986 & 93.886 & 35.76 & 87.99 & 93.39 & 91.93 & 30.1 & 80.81 & 88.09 & 86.19 & 40.91 & 91.6 & 95.58 & 94.49 \\
sinpos &  & 15.86 & 91.88 & 95.676 & 94.792 & 14.71 & 89.88 & 94.5 & 93.11 & 11.98 & 82.13 & 89.54 & 87.71 & 17.18 & 92.82 & 96.29 & 95.23 \\
CAPE, $\lambda=1$ &  & 51.032 & 91.96 & 95.834 & 94.988 & 48.04 & 89.77 & 94.48 & 93.02 & 40.55 & 82.73 & 89.58 & 88.14 & 54.47 & 93.07 & 96.43 & 95.3 \\
CAPE &  & 51.82 & 91.292 & 95.396 & 94.53 & 47.32 & 88.62 & 94.07 & 92.52 & 40.21 & 81.29 & 88.75 & 87.64 & 53.95 & 92.16 & 96.09 & 94.94 \\
\midrule
nopos+ft & \multirow{5}{*}{ \shortstack{ $224^2$ \\ $\downarrow$ \\ $384^2$} }  & 68.448 & 92.506 & 95.118 & 94.25 & 63.71 & 90.61 & 93.86 & 92.67 & 55.07 & 83.52 & 88.47 & 87.18 & 70.52 & 93.72 & 95.7 & 94.83 \\
abspos+ft &  & 54.056 & 93.958 & 95.96 & 95.112 & 49.48 & 92.58 & 95.11 & 93.6 & 42.31 & 86.42 & 90.19 & 88.39 & 56.36 & 95.12 & 96.85 & 95.8 \\
sinpos+ft &  & 42.796 & 93.574 & 96.078 & 95.194 & 38.91 & 91.85 & 95.08 & 93.8 & 33.17 & 85.54 & 90.03 & 88.28 & 44.75 & 94.76 & 96.98 & 95.68 \\
CAPE+ft, $\lambda=1$ &  & 79.264 & 94.146 & 96.14 & 95.47 & 75.63 & 92.86 & 95.04 & 94.02 & 66.43 & 86.84 & 90.62 & 89.13 & 81.36 & 95.39 & 96.94 & 95.96 \\
CAPE+ft &  & 79.992 & 93.818 & 96.06 & 95.322 & 76.09 & 92.31 & 95.24 & 94.16 & 67 & 85.91 & 90.04 & 88.75 & 81.7 & 95.02 & 96.97 & 96.09 \\
\midrule
UniViT, sinpos & \multirow{3}{*}{mix} &  94.216 &  95.4 &  96.036 &  95.964 &  92.93 &  94.28 &  94.93 &  94.76 &  86.55 &  88.8 &  89.98 &  89.93 &  95.28 &  96.31 &  96.73 &  96.65 \\
UniVit, CAPE, $\lambda=1$ &  &  94.392 &  95.558 &  96.176 &  96.016 &  93.02 & 94.48 &  95.23 &  95.19 &  86.72 &  88.91 &  90.3 &  90.33 &  95.29 &  96.4 &  97 &  96.84 \\
UniViT, CAPE &  &  94.354 &  95.438 & 96.044 &  95.716 &  92.92 &  94.36 &  95.11 &  94.69 &  86.76 &  89.18 &  90.45 &  89.68 &  95.5 &  96.42 &  96.87 &  96.37 \\
\midrule
abspos-S & $224^2$ & 92.126 & 94.692 & 94.094 & 85.75 & 90.78 & 93.74 & 92.97 & 81.28 & 83.44 & 87.62 & 86.72 & 73.72 & 93.82 & 95.95 & 95.3 & 85.73 \\
UniViT-S, CAPE, $\lambda=1$ & mix & 92.982 & 94.642 & 95.526 & 95.346 & 92.08 & 93.98 & 95.03 & 94.65 & 85.15 & 87.92 & 89.64 & 89.02 & 94.88 & 96 & 96.72 & 96.37 \\
\midrule
abspos-Ti & $224^2$ & 86.538 & 90.928 & 90.156 & 80.804 & 85.59 & 90.13 & 88.48 & 76.1 & 76.5 & 82.39 & 81.41 & 68.03 & 89.74 & 93.25 & 91.81 & 81 \\
UniViT-Ti, CAPE, $\lambda=1$ & mix & 86.942 & 89.738 & 91.332 & 90.798 & 86.03 & 88.79 & 90.35 & 89.25 & 77.33 & 80.97 & 83.42 & 82.41 & 90.17 & 92.31 & 93.44 & 92.37 \\
\bottomrule
\end{tabular}
}
\end{center}
\end{table}

For all ViT/UniViT models presented in Figures~\ref{fig:vision:vit_main} and~\ref{fig:vision:mix_sizes}, and in the ablation study below, we report their top-1 and top-5 accuracies in Tables~\ref{tab:vision_all_top1} and~\ref{tab:vision_all_top5}, respectively, evaluated on the ImageNet validation and ImageNet-v2 test sets and on images with different resolutions. 

We train models in Flashlight framework\footnote{\scriptsize\texttt{https://github.com/flashlight/flashlight}} where ViT/DeiT~\cite{touvron2020training} training is reproduced following an original implementation:\footnote{\scriptsize\texttt{https://github.com/facebookresearch/deit}} initialization is set to the truncated normal distribution, Rand-Augment~\cite{cubuk2020randaugment}, Mixup~\cite{zhang2017mixup} and Cutmix~\cite{yun2019cutmix}, random erasing~\cite{zhong2020random} and repeated augmentation~\cite{hoffer2020augment,berman2019multigrain} are used as data augmentations; training is done with AdamW optimizer for 300 epochs.
All models use learnable absolute positional embedding for the class token. 
We train ViT (UniViT) models on 16 GPUs, V100 32GB, with mixed-precision and batch size 64 per GPU for 19-56h (37h) depending on the input resolution ($384^2$ resolution is trained on 32 GPUs with batch size 32 per GPU).  

In Figure~\ref{fig:vision:speed} the throughput is measured as the number of images that can be processed per second on single 16GB V100 GPU following the benchmarking method from~\cite{touvron2020training}: 
for each image resolution we pass the largest possible batch size and calculate the average time over 30 runs to process that batch.

\subsection{Finding the Best Resolution for UniViT Evaluation}\label{app:vision:resize}

In this section we describe an evaluation procedure that improves UniViT with CAPE performance by resizing input images to an optimal resolution. 
We split ImageNet validation images into 8 bins, according to their size $s=\min(h, w)$, where $h$ and $w$ are image height and width, respectively: $s\in[54,100]$, $s\in[101,150]$, $s\in[151,200]$, $s\in[201,250]$, $s\in[251,300]$, $s\in[301,350]$, $s\in[351,384]$, $s\in[385, \inf]$. 
For each bin we consider several resizing strategies: 
i) resize all images in a bin either to $160^2$, or $224^2$, or $384^2$; 
ii) resize all images to the minimum $s$ value in a bin, \textit{Min}; 
iii) resize all images to the maximum $s$ value in a bin, \textit{Max}; 
iv) use image's original size but still perform a central \textit{rectangular} crop in a similar manner as standard evaluation is done for ImageNet, \textit{Original}; 
v) use image's original size, \textit{Original (no crop)}. 
For the bin with high resolution images $[385, \inf]$ we use $500^2$ as the maximum resize value and apply neither iv) nor v) strategies as there are images with $s$ > 5000 px.
We report top-1 accuracy for this evaluation procedure with different strategies per each bin in Table~\ref{tab:vision:own_size_eval}: for the best strategy in each bin (table row) we report accuracy (\%) while for other strategies in the same bin we report absolute drop in accuracy compared to the best value. 
Best values are additionally marked in bold.

\begin{table}[h!]
\caption{UniViT with CAPE evaluation on ImageNet validation set with different strategies on input resizing. Images are split into 8 bins by minimum spatial size. 
We report best top-1 accuracy (\%) in each row (bold) and the drop in accuracy compared to this best accuracy for other columns.\label{tab:vision:own_size_eval}}
\begin{center}
\resizebox{0.9\linewidth}{!}{
\begin{tabular}{@{}cccccccccc@{}}
\toprule
 \# Images & Min Res. & Max Res. & $160^2$ & $224^2$ & $384^2$ & Min & Max & Original & Original (no crop) \\
\midrule
146 & 54 & 100 & \textbf{84.25} & -0.69 & -0.69 & -19.86 & -1.37 & -6.16 & -8.22 \\
221 & 101 & 150 & -0.81 & -1.36 & -0.45 & -5.43 & -2.71 & -5.43 & \textbf{74.66} \\ 
372 & 151 & 200 & -1.61 & -1.08 & -2.42 & -3.23 & -2.42 & \textbf{78.49} & -1.61 \\
538 & 201 & 250 & -3.16 & -1.30 & -1.49 & -2.60 & -1.86 & \textbf{81.97} & -2.42 \\
1090 & 251 & 300 & -4.40 & -1.47 & -0.28 & -1.56 & -1.01 & \textbf{79.72} & -0.37 \\
8496 & 301 & 350 & -3.83 & -1.88 & -0.48 & -1.39 & -1.08 & \textbf{83.57} & -0.48 \\
24538 & 351 & 384 & -4.14 & -1.88 & -0.43 & -0.79 & -0.43 & \textbf{82.10} & -0.13 \\
14599 & 385 & - & -3.30 & -1.25 & -0.14 & -0.13 & \textbf{84.46} & - & - \\ 
\bottomrule
\end{tabular}
}
\end{center}
\end{table}

\subsection{Visualization of Positional Embeddings}

\begin{figure}[h!]
\begin{subfigure}{.2\textwidth}
  \centering
  \includegraphics[width=0.65\textwidth]{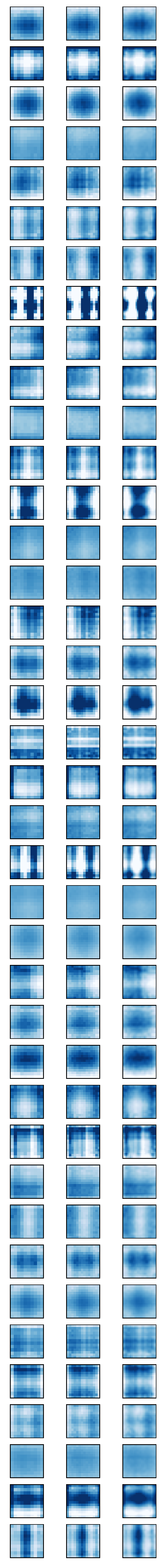}
\end{subfigure}
\begin{subfigure}{.2\textwidth}
  \centering
  \includegraphics[width=0.65\textwidth]{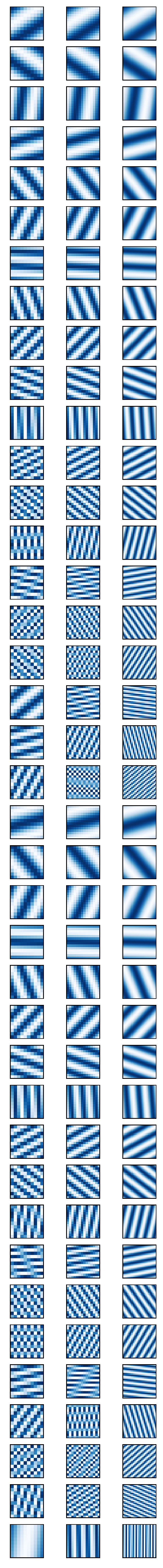}
\end{subfigure}
\begin{subfigure}{.2\textwidth}
  \centering
  \includegraphics[width=0.65\textwidth]{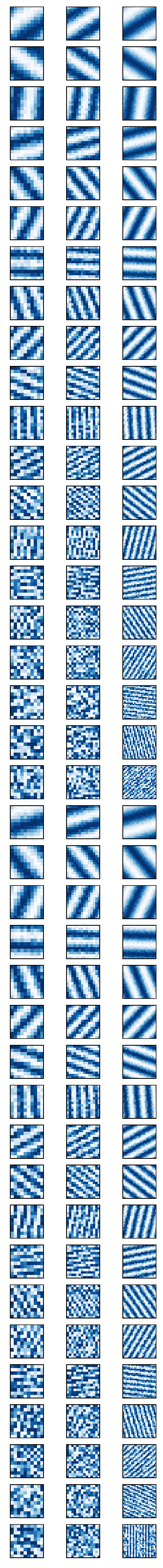}
\end{subfigure}
\caption{Visualization of positional embeddings for ViT models trained on $224^2$ resolution: \textit{abspos} (left), \textit{sinpos} (middle) and CAPE (right).
Each column consists of 3 sub-columns corresponding to input resolutions $160^2$, $224^2$ and $384^2$. 
Only some components (each 20th out of 768) are shown.
When \textit{sinpos} applied to low-resolution images, spacial aliasing is visible in latest components of embeddings. 
CAPE's augmentations destruct this patterns and prevent model from over-fitting.
\label{fig:emb_visualization}}
\end{figure}

We visualize positional embeddings (excluding class token) for ViT models trained on $224^2$ input resolution in Figure~\ref{fig:emb_visualization}: 
for each positional embedding we plot every 20th among 768 components (each row corresponds to a particular component) and visualize its values with the image of shape $(r/16)^2$ where $r^2$ is an input image resolution. 
We consider input resolutions $160^2$, $224^2$ and $384^2$ shown as left, middle and right sub-columns for each positional embedding in Figure~\ref{fig:emb_visualization}.

\subsection{Ablations}\label{app:vision:ablation}

Both \textit{sinpos} and CAPE first re-scale patch positions $(x, y)$ to the $[-1, 1]$ interval independently from the image resolution. 
We study if an alternative re-scaling of patch positions during inference improves performance on resolutions other than training ones. 
For ViT models trained with either \textit{sinpos} or CAPE on $224^2$ resolution we perform evaluation on $r^2=160^2$ and $r^2=384^2$ resolutions by re-scaling $(x, y)$ to $[-\gamma, \gamma]$: $\gamma$ is set to either $1$ (baseline strategy), $r/224$, or $\sqrt{(r/224)}$. 
Results of this comparison, Table~\ref{tab:abl:vision:scaling}, are consistent across models and suggest that for applying to smaller resolution ($160^2$) decreasing scale $\gamma$ to match density of patches on a plane to train-time is beneficial; 
however, opposite effect is observed when model is applied to higher resolution ($384^2$) inputs, potentially because distances between patch positions in this case were not observed at training time.
For simplicity we use re-scaling to $[-1, 1]$ in the rest of our experiments.

\begin{table}[h!]
\begin{center}
\caption{Ablation study on re-scaling positions to the range $[-\gamma, \gamma]$ for ViT models trained on $224^2$ images with \textit{sinpos} or CAPE. We report top-1 accuracy (\%) on ImageNet validation set evaluated on images scaled to $160^2$ and $384^2$ resolutions.\label{tab:abl:vision:scaling}}
\resizebox{0.5\linewidth}{!}{
\begin{tabular}{@{}cccc@{}}
\toprule
Model & $\gamma$ & Top-1, $r=160$ & Top-1, $r=384$ \\
\midrule
\multirow{3}{*}{sinpos} & $r/224$ & 77.89 & 72.01 \\
 & $\sqrt{r/224}$ & 77.11 & 76.94 \\
 & 1 & 77.15 & 79.72 \\
\midrule
\multirow{3}{*}{CAPE, $\lambda=1$} & $r/224$ & 77.96 & 80.18 \\
 & $\sqrt{r/224}$ & 77.80 & 80.51 \\
 & 1 & 77.70 & 80.38 \\
\midrule
\multirow{3}{*}{CAPE} & $r/224$ & 77.28 & 80.20 \\
 & $\sqrt{r/224}$ & 77.18 & 80.28 \\
 & 1 & 77.14 & 80.33 \\

\bottomrule
\end{tabular}
}
\end{center}
\end{table}

\begin{figure}[htb!]
  \centering
  \includegraphics[width=0.9\textwidth]{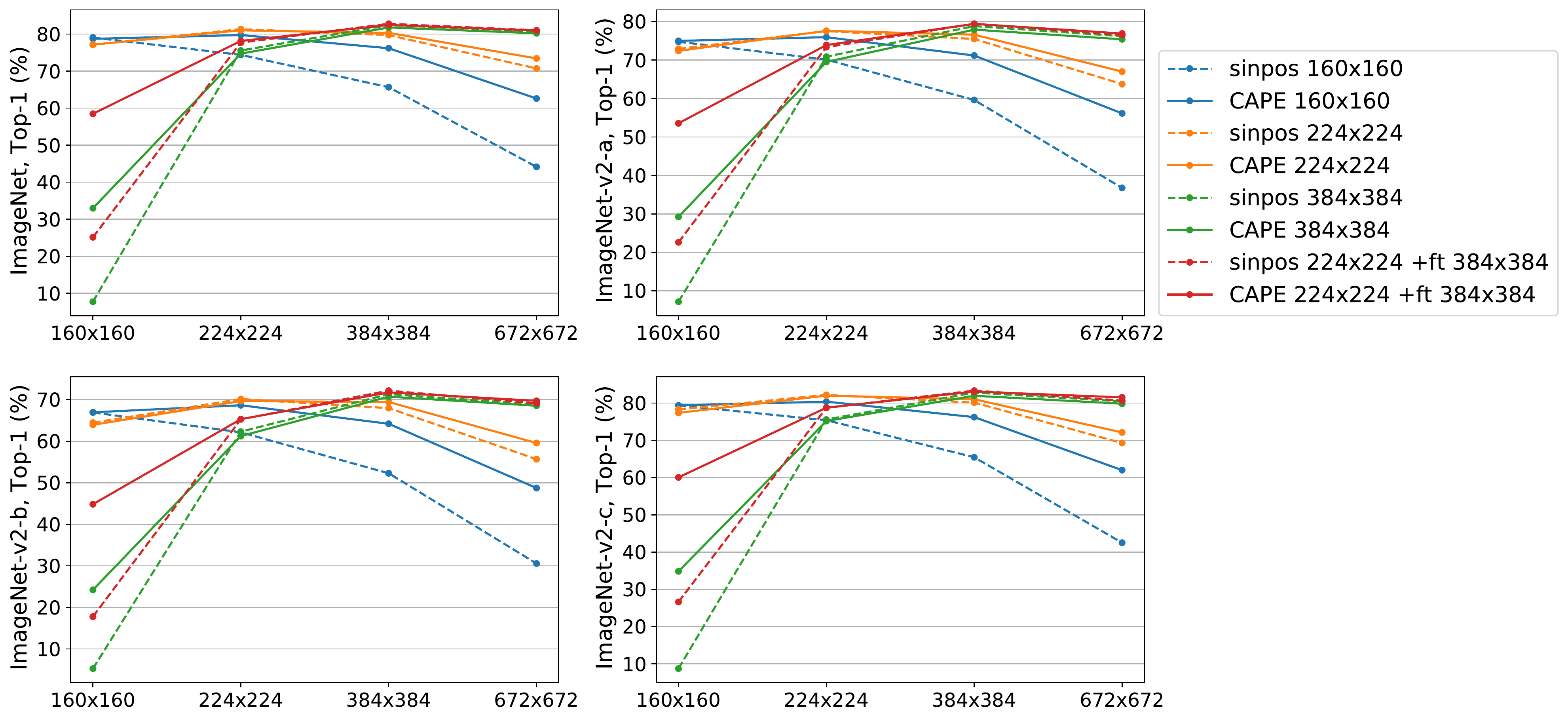}
  \caption{Comparison of top-1 accuracy between \textit{sinpos} and CAPE trained on either $160^2$, or $224^2$, or $384^2$ resolutions and evaluated across the board. 
  Models trained on $224^2$ resolution and further fine-tuned on $384^2$ resolution are marked with ``+ft''.\label{fig:abl:vis_sin_vs_cape}}
\end{figure}

\begin{figure}[t!]
  \centering
  \includegraphics[width=0.9\textwidth]{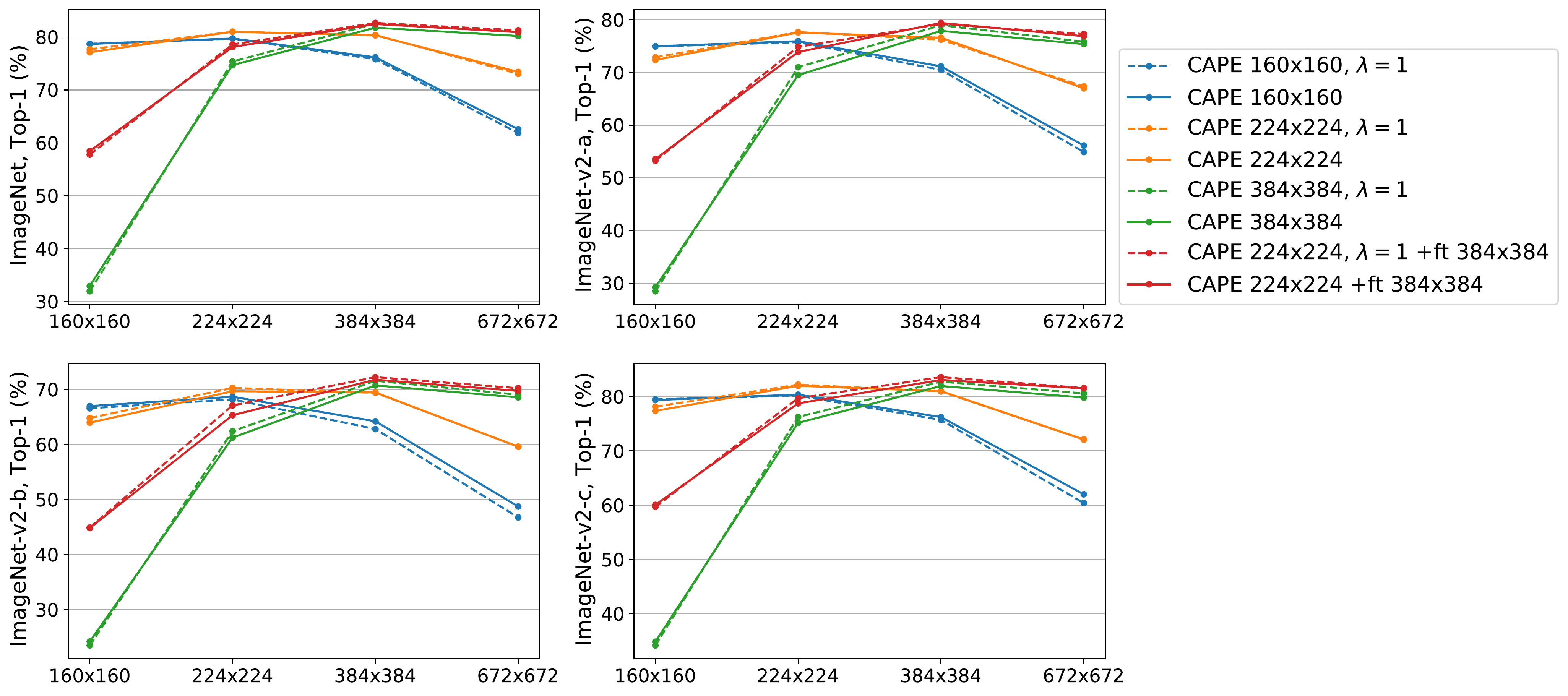}
  \caption{Comparison of top-1 accuracy for CAPE with $\lambda_{max}=1$ (dashed) and $\lambda_{max}=1.4$ (solid) trained on either $160^2$, or $224^2$, or $384^2$ resolutions and evaluated across the board. 
  Models trained on $224^2$ resolution and further fine-tuned on $384^2$ resolution are marked with ``+ft''.\label{fig:abl:vis_scale_vs_no_scale_cape}}
\end{figure}

In Figure~\ref{fig:abl:vis_sin_vs_cape}, we compare \textit{sinpos} and CAPE for ViT models in more detail. 
Overall, CAPE performs better or similar to \textit{sinpos} on the training resolution while it significantly outperforms \textit{sinpos} on other resolutions. In Figure~\ref{fig:abl:vis_scale_vs_no_scale_cape} we study the importance of scale $\lambda$ for CAPE in ViT models. 
Scale $\lambda_{max} > 1$ slightly improves generalization for higher and lower resolutions.

\begin{figure}[tb!]
  \centering
  \includegraphics[width=0.85\textwidth]{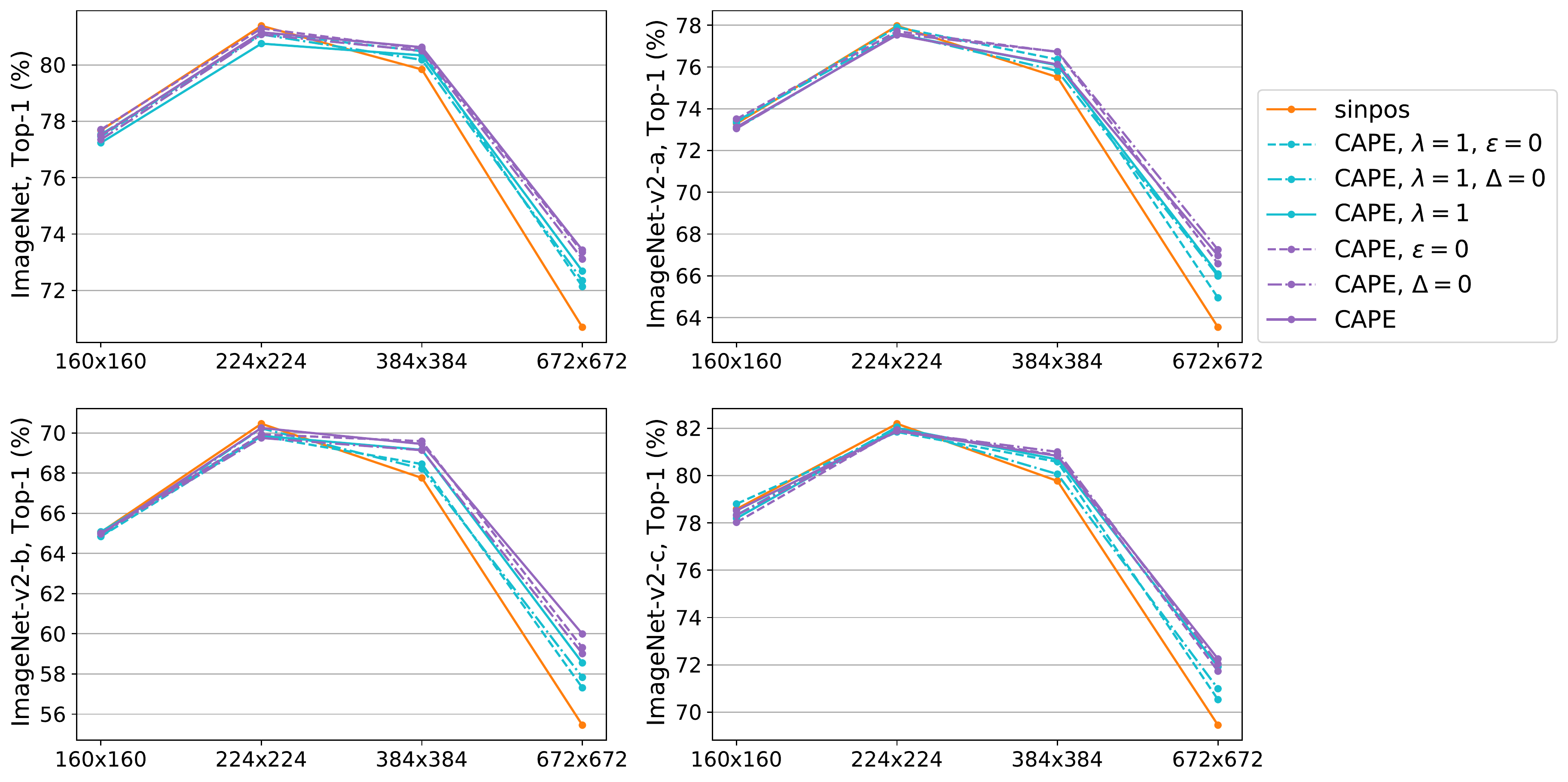}
  \caption{
      Comparison of top-1 accuracy between \textit{sinpos} and CAPE with different configurations on global, local shifts and global scaling trained on $224^2$ resolution.\label{fig:abl:vis_ablations}
  }
\end{figure}

In Figure~\ref{fig:abl:vis_ablations} we study the importance of global $\Delta$ and local $\epsilon$ shifts for CAPE in ViT models trained on $224^2$ resolution. On higher resolutions, $384^2$ and $672^2$, models with both shifts (solid) perform similar or better than models trained with either local (dotted-dashed) or global (dashed) shifts. Overall, only one of the shifts, global or local, can be used while the most important CAPE's parameter is the global scale. On the other hand, any combination of augmentations in CAPE clearly outperforms \textit{sinpos} on resolutions different from training one.

\begin{figure}[tbh!]
  \centering
  \includegraphics[width=0.85\textwidth]{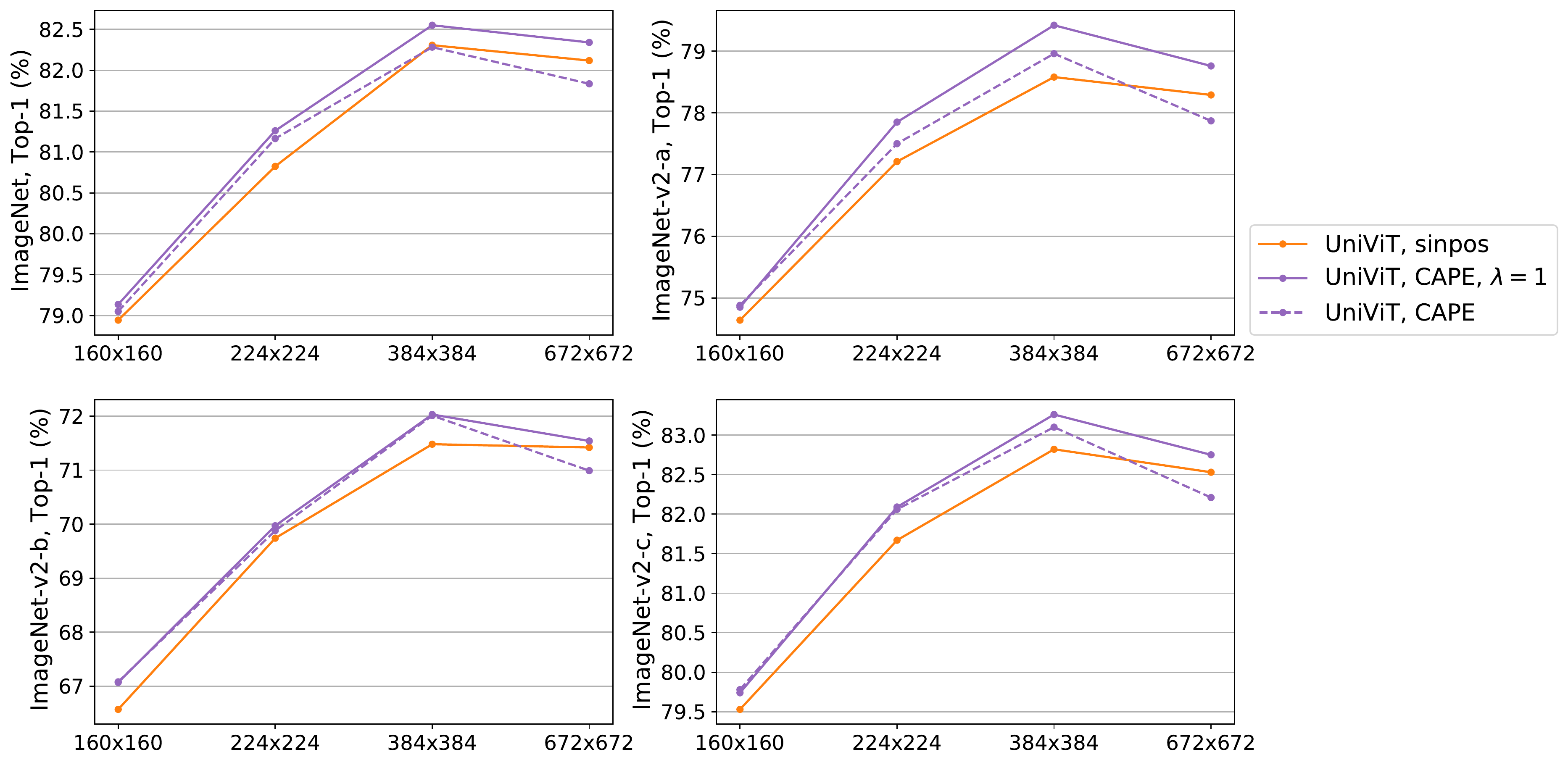}
  \caption{
      Comparison of top-1 accuracy between \textit{sinpos} and CAPE (with and without global scaling) trained on the mixture of resolutions 
      $\{128^2, 160^2, 192^2, 224^2, 256^2, 288^2, 320^2\}$.\label{fig:abl:vis_univit}
  }
\end{figure}

In Figure~\ref{fig:abl:vis_univit} we study if CAPE's augmentations are beneficial for UniViT model, compared to using UniViT with \textit{sinpos}.
Overall \textit{sinpos} performs worst among UniViT models, while outperforming UniViT with CAPE and global scaling on $672^2$ resolution. 
UniViT with CAPE and no global scaling ($\lambda_{max}=1$) performs the best on all resolutions, suggesting that variability in training resolutions provides a sufficient base for generalization to higher resolutions.

\newpage

\section{Automatic Speech Recognition Experiments}

\subsection{Data}\label{app:asr:data}
For WSJ data we consider the standard subsets \emph{si284}, \emph{nov93dev} and \emph{nov92} for training, validation and test, respectively. 
We remove any punctuation tokens from \emph{si284} transcriptions before training. 
TED-LIUM v3 dataset is based on TED conference videos. 
We use the last edition of the training set (v3); validation and test sets are kept consistent (and thus numbers are comparable) with the earlier releases. 
We follow the Kaldi recipe~\cite{daniel2011kaldi} for data preparation.
In Tables~\ref{tab:data_asr_stat} and~\ref{tab:data_asr} we present statistics of the datasets used in Section~\ref{sec:asr}. One could notice that TED-LIUM v3 validation and test sets have samples with significantly longer duration and larger number of words in their transcriptions, which makes these sets the most challenging among other public data. 

\begin{table}[h!]
\caption{Statistics on datasets: sampling frequency, duration (in hours), and speech type. \label{tab:data_asr_stat}}
\begin{center}
\begin{tabular}{@{}cccccc@{}}
\toprule
Data & kHz & Train (h) & Valid (h) & Test (h) & Speech \\
\midrule
WSJ & 16 & 81.5 & 1.1 & 0.7 & read \\
TL & 16 & 452 & 1.6 & 2.6 & oratory \\
LS & 16 & - & - & 5.4+5.4 & read \\
RV & 16 & - & - & 18.8+19.5 & diverse \\
\bottomrule
\end{tabular}
\end{center}
\end{table}


\begin{table}[h!]
\caption{Statistics on datasets: mean sample duration (in seconds) and mean sample transcription length (in words). \label{tab:data_asr}}
\begin{center}
\resizebox{\linewidth}{!}{
\begin{tabular}{@{}cccccccc@{}}
\toprule
Data & Train $\mu\pm\sigma$ (s) & Valid $\mu\pm\sigma$ (s) & Test $\mu\pm\sigma$ (s) & Train $\mu\pm\sigma$ (wrd) & Valid $\mu\pm\sigma$ (wrd) & Test $\mu\pm\sigma$ (wrd) \\
\midrule
WSJ & $7.8 \pm 2.9$ & $7.8 \pm 2.9$ & $7.6 \pm 2.5$ & $17 \pm 7$ & $16\pm 7$ & $17\pm6$ \\
TL & $6 \pm 3$ & $11.3 \pm 5.7$ & $8.1 \pm 4.3$ & $17 \pm 10$ & $35\pm 20$ & $24\pm 15$ \\
LS & - & - & $7 \pm 4.8$ & - & - & $19\pm13$ \\
\bottomrule
\end{tabular}
}
\end{center}
\end{table}

\subsection{Acoustic Model Training}\label{app:asr:am}
For all experiments we compute 80 log-mel spectrogram features for a 25ms sliding window, strided by 10ms (unless we explicitly vary STFT hop distance). 
All features are normalized to have zero mean and unit variance per input sequence before feeding into the neural network. 

The self-attention dimension is $768$ and the feed-forward network (FFN) dimension is $3072$ in each Transformer layer. We use dropout $0.3$ after the convolution layer; for all Transformer layers, we use dropout on the self-attention and on the FFN, and layer drop~\cite{fan2019reducing}, dropping entire layers at the FFN level.
Transformer dropout and layer drop values are set to be $0.4$ for WSJ and $0.1$ for TED-LIUM v3 training.

SpecAugment~\cite{park2019specaug} is used for data augmentation during training: there are two frequency masks, and ten time masks with maximum time mask ratio of $p=0.1$, the maximum frequency bands masked by one frequency mask is 30, and the maximum frames masked by the time mask is 50; time warping is not used. 
We use the Adagrad optimizer~\cite{duchi2011adaptive}. 
All models are trained with dynamic batching (average batch size is 240s/GPU) and mixed-precision computations on 16 GPUs (Volta 32GB) for 1 day on WSJ and 3-4 days on TED-LIUM v3. 
All ASR experiments are done within Flashlight framework on top of the publicly available training configurations\footnote{\scriptsize\texttt{https://github.com/flashlight/wav2letter/tree/master/recipes/rasr}} for baselines with \textit{relpos} from~\cite{likhomanenko2020rethinking}.

\begin{figure}[t!]
\begin{subfigure}{.48\textwidth}
  \centering
  \includegraphics[width=0.7\textwidth]{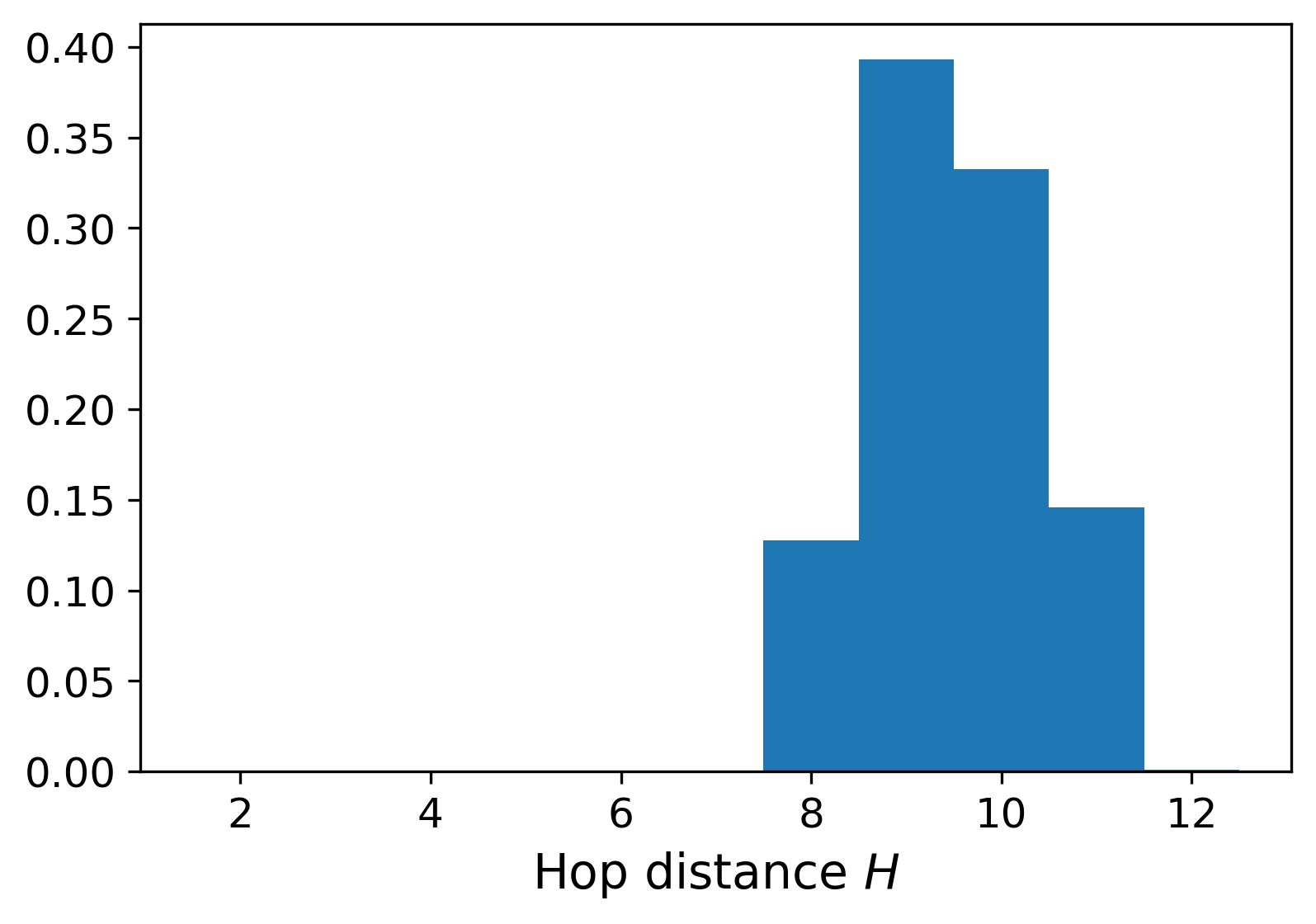}
\end{subfigure}
\begin{subfigure}{.48\textwidth}
  \centering
  \includegraphics[width=0.7\textwidth]{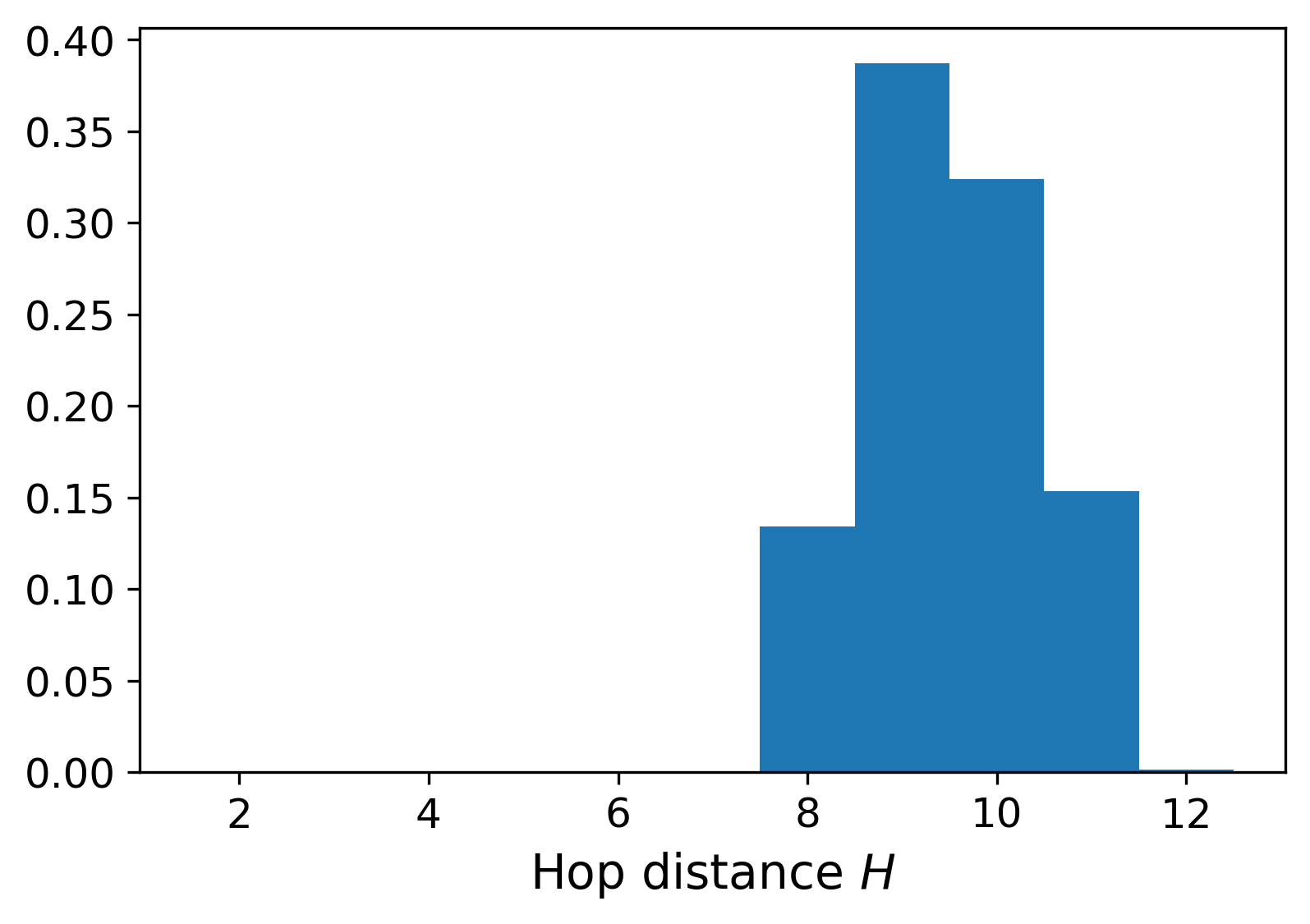}
\end{subfigure}
\caption{Hop distance distribution for WSJ (left) and TED-LIUM v3 (right) data. \label{fig:abl:speech:hop_pdf}}
\end{figure}

\subsection{Padding-free ASR with CAPE and Variable STFT Hop Distance}\label{app:asr:hop}

\begin{figure}[h!]
  \centering
  \includegraphics[width=0.9\textwidth]{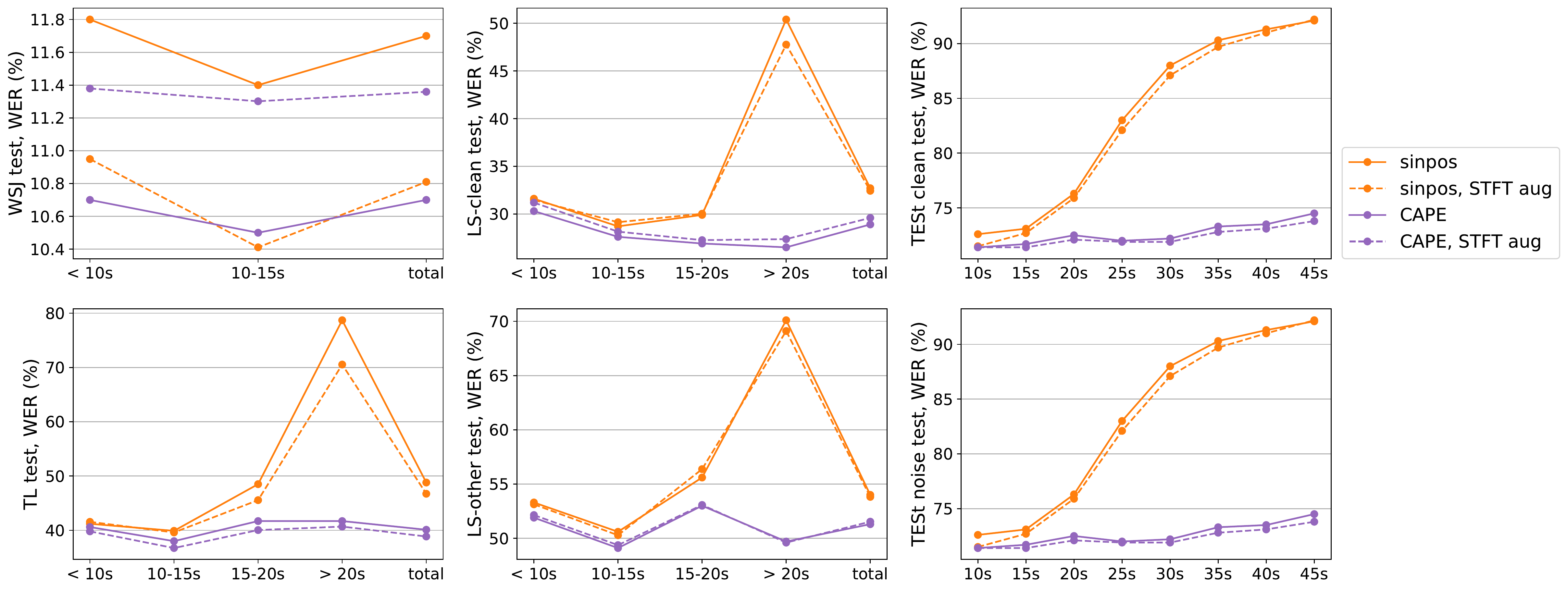}
  \caption{Word error rate comparison for models trained on WSJ data with \textit{sinpos} or CAPE ($\lambda=1$) with classical pipeline (solid) or with variable STFT hop distance (dashed).\label{fig:abl:wsj_hop}}
\end{figure}

\begin{figure}[h!]
  \centering
  \includegraphics[width=0.9\textwidth]{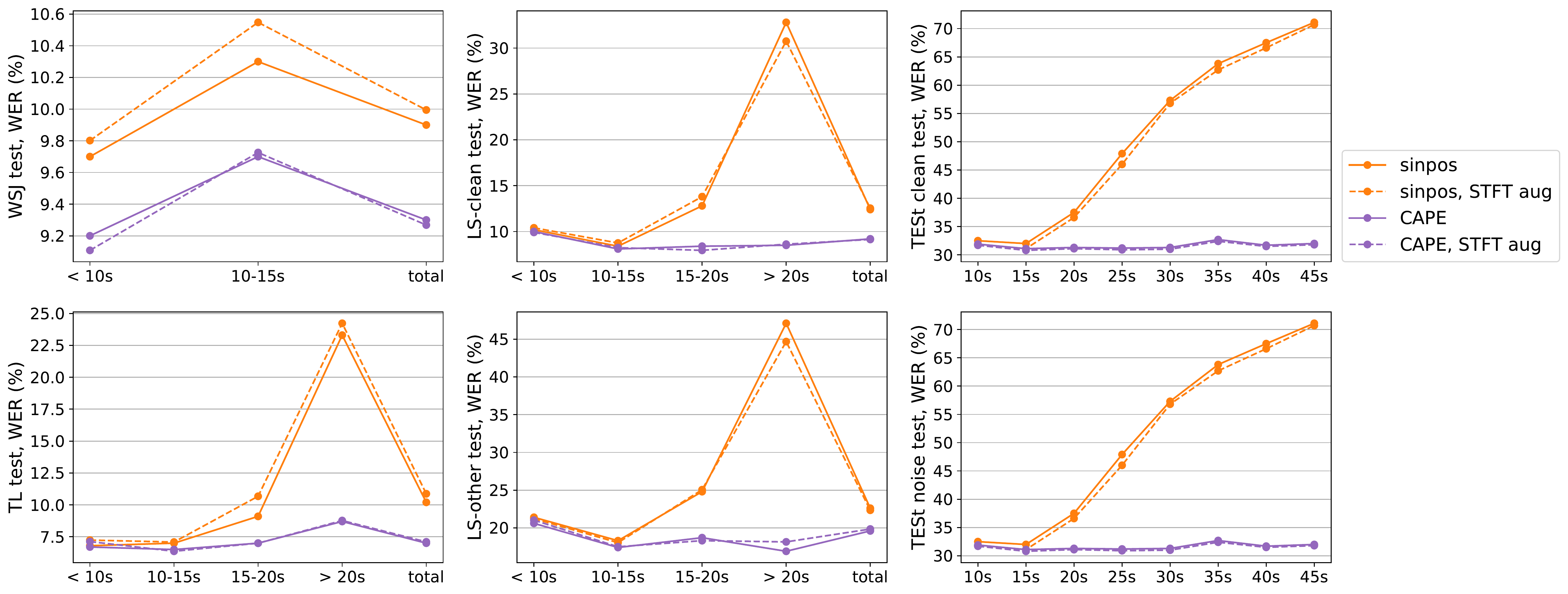}
  \caption{Word error rate comparison for models trained on TED-LIUM v3 data with \textit{sinpos} or CAPE ($\lambda=1$) with classical pipeline (solid) or with variable STFT hop distance (dashed).\label{fig:abl:tl_hop}}
\end{figure}

We have implemented pipeline where padding is no longer used to form a batch from samples with different input duration. 
For each audio in the batch short-time Fourier Transform (STFT) hop distance $H$ is set in a way that output number of frames is the same (except rounding) as for hypothetical audio which has duration equal to the mean over batch and is processed with $H=10$ms.
Because the hop distance is an integer number, number of frames after STFT is matched only approximately within a batch, so we reduce the number of frames in each sample to match the shortest sample in the batch by randomly and uniformly skipping frames. 
To have low variation of samples duration in a batch (which implies limited vatiation in $H$) the following shuffling strategy is performed for every epoch: 
i) compute perturbed sample duration by multiplying original sample duration by a random number from $\mathcal{U}(0.85, 1.15)$; 
ii) sort samples by their perturbed duration;
iii) batches are formed by grouping sequential samples. 
Example of hop distance distribution after proposed shuffling strategy for WSJ and TL data is shown in Figure~\ref{fig:abl:speech:hop_pdf}.
For both \textit{sinpos} and CAPE embeddings we train models with this new pipeline and observe mostly lower WER and improved generalization, especially on TL test and RV data which are the most challenging among evaluation sets, Figures~\ref{fig:abl:wsj_hop} and~\ref{fig:abl:tl_hop}.\footnote{For all models evaluation the batch size is set to 1 and $H=10$ms, thus padding never affects the performance on validation and test sets.}

\begin{figure}[h!]
  \centering
  \includegraphics[width=0.45\textwidth]{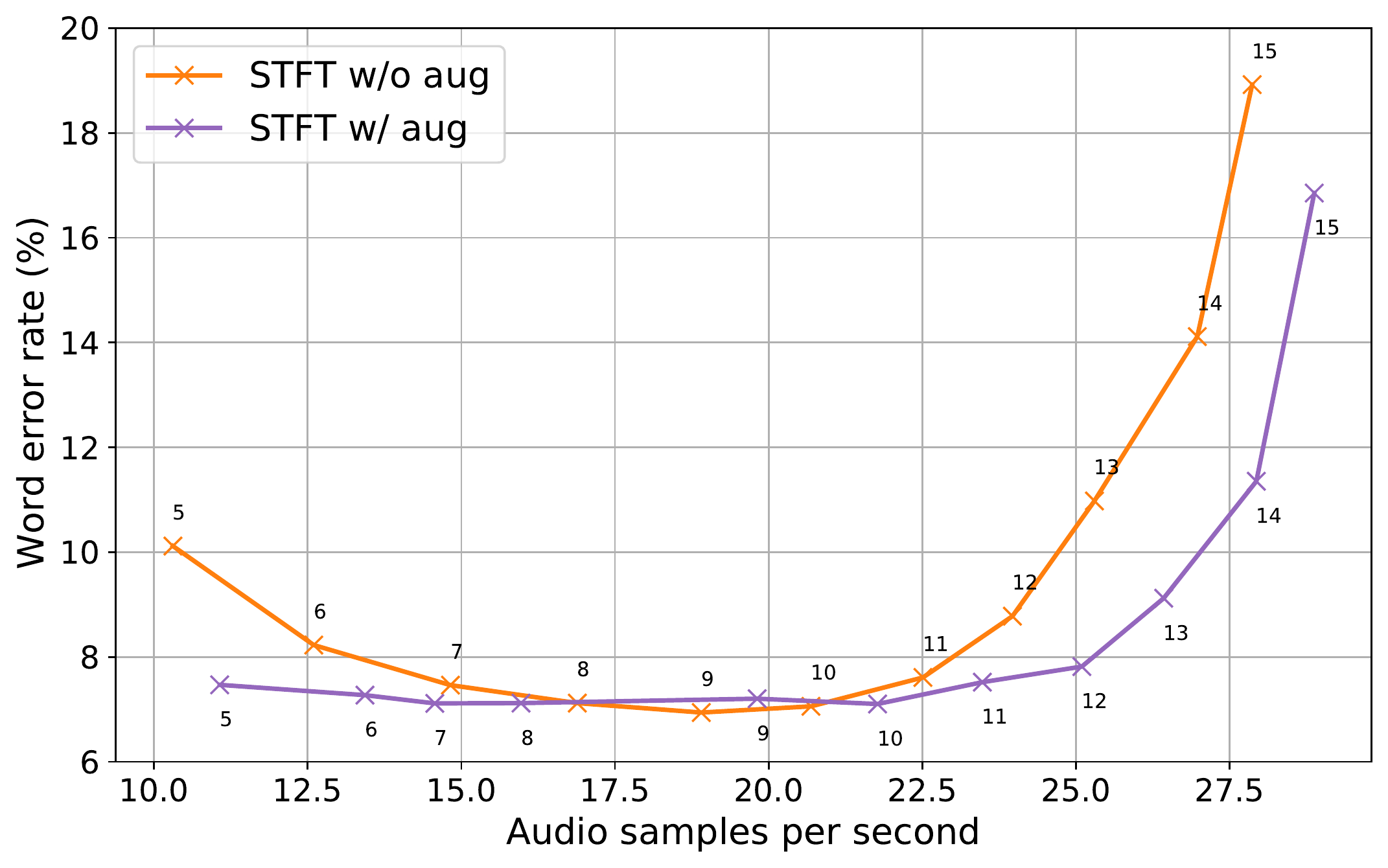}
  \caption{Dependence between throughput and word error rate on TED-LIUM v3 test set for models trained with CAPE ($\lambda=1)$ on TED-LIUM v3 data: with classical pipeline (orange) and with variable STFT hop distance (purple). Different throughput values correspond to different STFT hop distances in ms (crosses) used during evaluation.\label{fig:abl:tl_hop_time}}
\end{figure}

As discussed in Section~\ref{sec:asr}, STFT hop distance can be viewed as image resolution in vision and padding-free ASR – as UniViT. By adjusting hop distance during inference for padding-free ASR we can achieve higher throughput with similar recognition quality, see Figure~\ref{fig:abl:tl_hop_time}. Moreover, it is robust to STFT hop distance variation having almost the same WER for $H\in[5, 12]$ms and less performance degradation for $H>12$ms.

\subsection{Ablations}\label{app:asr:ablation}

\begin{figure}[ht!]
  \centering
  \includegraphics[width=0.9\textwidth]{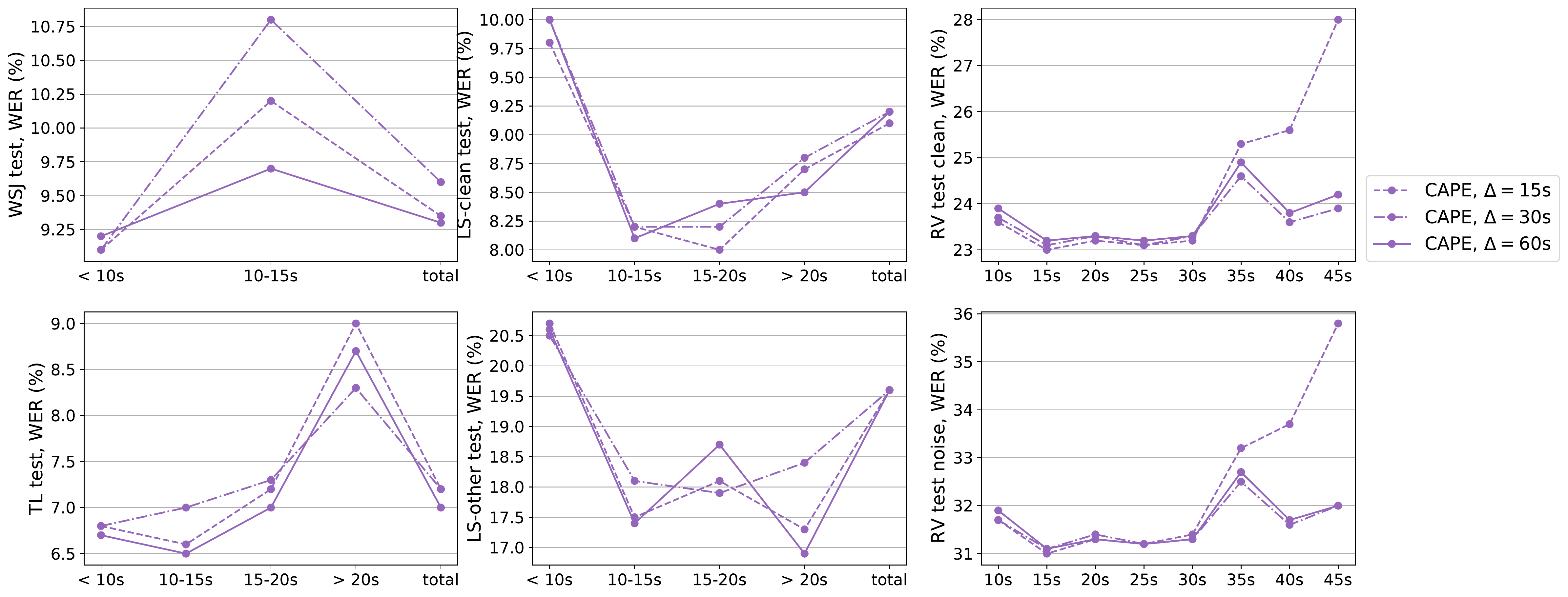}
  \caption{Word error rate comparison for models trained on TED-LIUM v3 data with CAPE and different global shifts which cover 15, 30 or 60s. The global scale is set to $\lambda_{max} =1$.\label{fig:abl:speech:global_shift_tl}}
\end{figure}

First, we study dependence between the global shift value $\Delta_{max}$ and model's performance and generalization abilities to the long duration. 
Varying the global shift we observe in Figure~\ref{fig:abl:speech:global_shift_tl} that larger global shift leads to a  better generalization on longer audios, so that CAPE with 30-60s global shifts is able to process 45s audio with the same performance as 10s on RV data.

\begin{figure}[htbp!]
  \centering
  \includegraphics[width=0.9\textwidth]{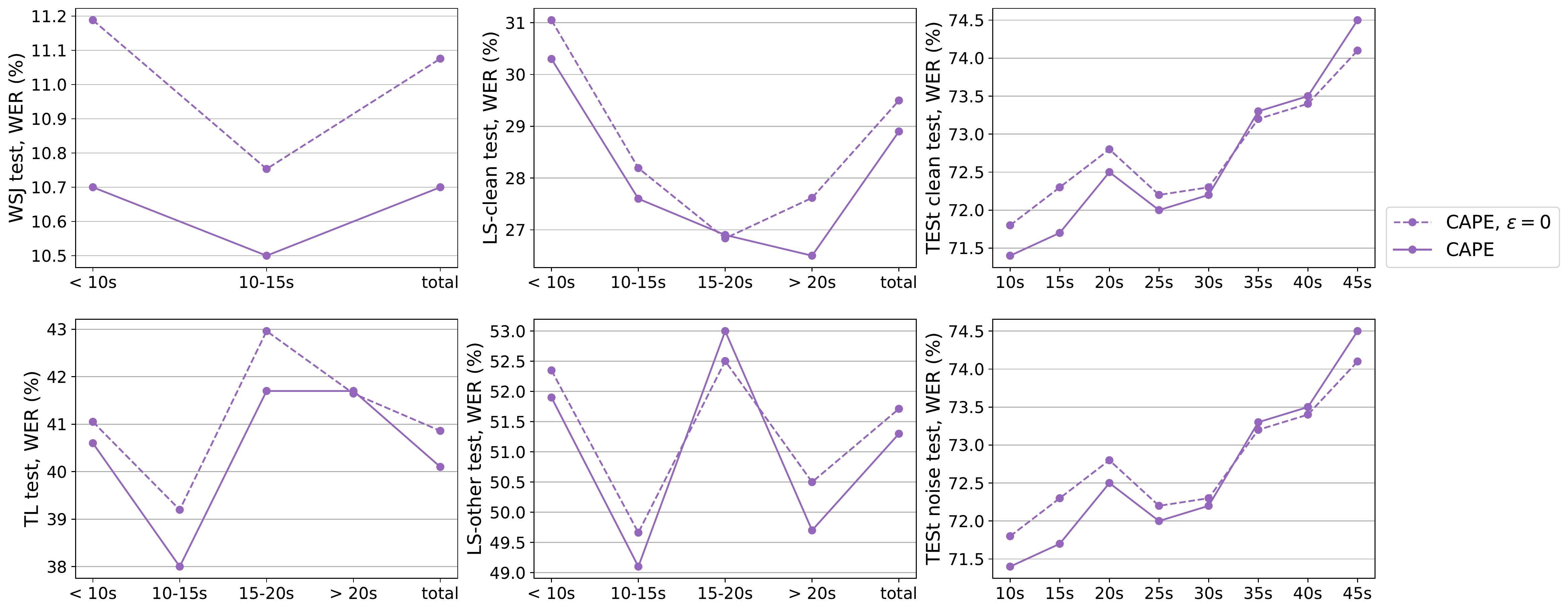}
  \caption{Word error rate comparison for CAPE models trained on WSJ data with global shift only (solid) or with global and local shifts together (dashed). The global scale is set to be $\lambda_{max} =1$.\label{fig:abl:speech:asr_wsj_local}}
\end{figure}

Secondly, we study the necessity of the local shift in CAPE. 
In Figure~\ref{fig:abl:speech:asr_wsj_local} we observe that local shift absence hurts the performance and generalization across the board.

\begin{figure}[t!]
  \centering
  \includegraphics[width=0.9\textwidth]{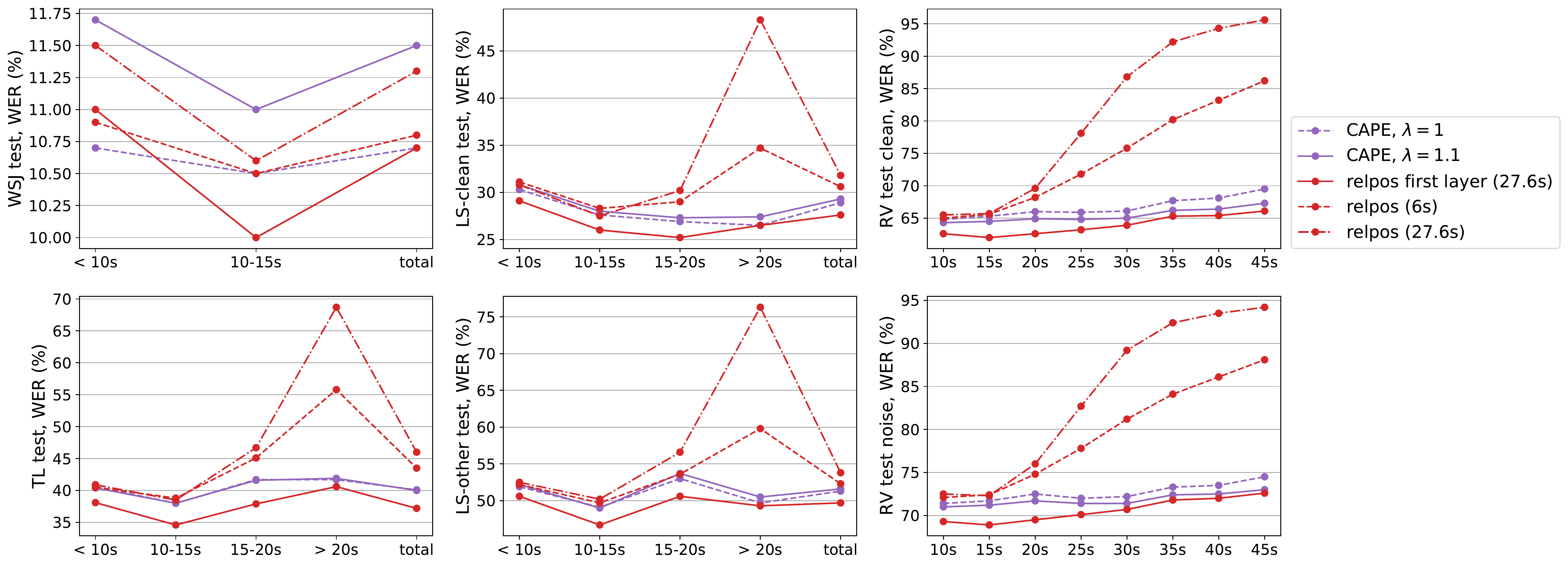}
  \caption{Word error rate comparison for models trained on WSJ data with different positional embeddings. 
  \textit{Relpos first layer} refers to a model where \textit{relpos} is used only in the first Transformer layer with 27.6s context to the left/right and no other positional embeddings are used.
  \label{fig:abl:speech:relpos_scale_wsj}}
\end{figure}

\begin{figure}[t!]
  \centering
  \includegraphics[width=0.9\textwidth]{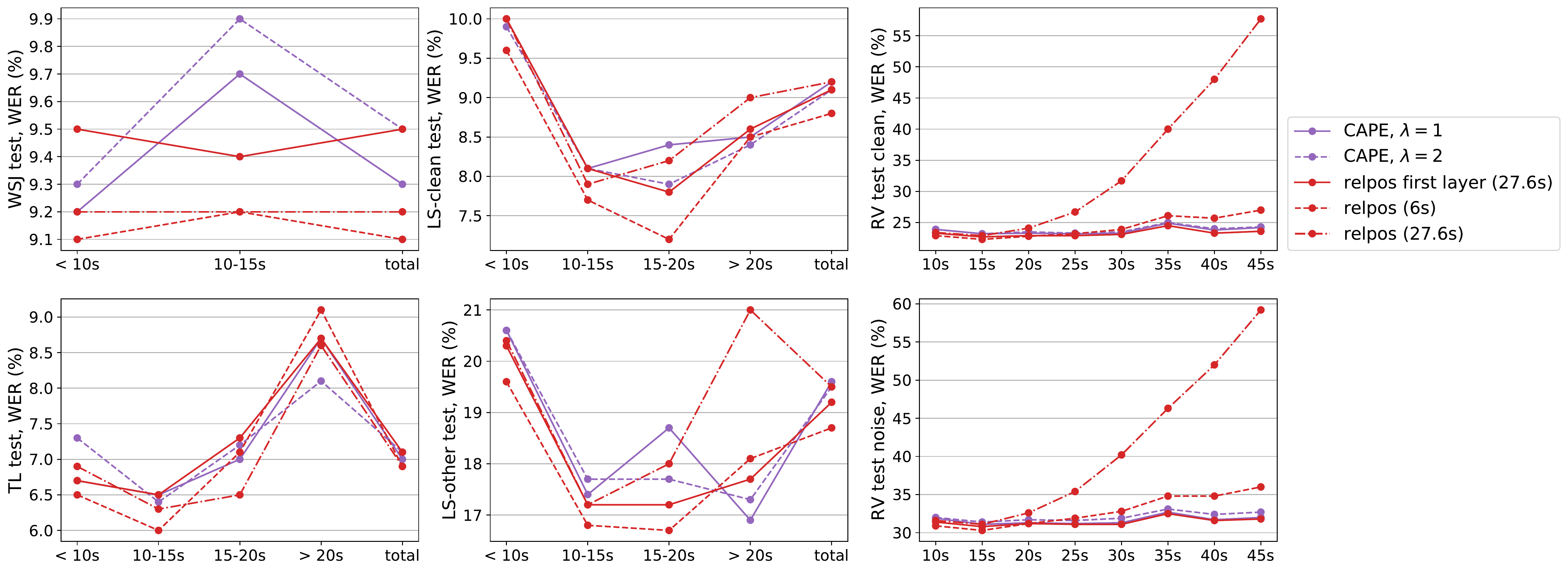}
  \caption{Word error rate comparison for models trained on TED-LIUM v3 data with different positional embeddings. 
  \textit{Relpos first layer} refers to a model where \textit{relpos} is used only in the first Transformer layer with 27.6s context to the left/right and no other positional embeddings are used.
  \label{fig:abl:speech:relpos_scale_tl}}
\end{figure}

Thirdly, we study the necessity of the global scaling in CAPE. 
In Figure~\ref{fig:abl:speech:relpos_scale_wsj} we observe that for WSJ models global scaling hurts a bit performance on public data while performs and generalizes better for RV data. 
In contrast, for TL models we observe in Figure~\ref{fig:abl:speech:relpos_scale_tl} that overall the global scaling improves performance on public data while hurts performance on RV data. Thus, the global scaling should be tuned separately depending on the data type.

As an ablation study we perform additional experiments with \textit{relpos}. 
First, we restrict \textit{relpos} context to small duration, 6s, to prevent over-fitting to relative positions: \textit{relpos 6s} outperforms \textit{relpos} on both public and RV data for both models trained on WSJ and TL having significantly better generalization to long audio durations. 
\textit{Relpos 6s} is performing similar to \textit{abspos} for duration $>20$s while CAPE still outperforms \textit{relpos 6s} on $>20$s, Figures~\ref{fig:abl:speech:relpos_scale_wsj} and~\ref{fig:abl:speech:relpos_scale_tl}. Independently, we also tried sinusoidal relative positional embedding~\cite{dai2019transformer} but it performed worse than learnable relative positional embedding.

Second, having in mind that \textit{nopos} performs well for a model trained on TL and CAPE's ability to learn spatial relations using a single layer, we wonder if \textit{relpos} should be used only in the first Transformer layer (in literature \textit{relpos}, when used, is applied in every attention layer). 
We modify \textit{nopos} model by injecting \textit{relpos} embedding (27.6s context to the right/left) only in the first Transformer layer: no any other Transformer layers use any positional embeddings, Figures~\ref{fig:abl:speech:relpos_scale_wsj} and~\ref{fig:abl:speech:relpos_scale_tl}. 
This \textit{relpos first layer} model behaves surprisingly well: for a WSJ model it outperforms CAPE on both public and RV data; for a TL model it behaves similar to CAPE on public data and a bit better on RV data. 
Both CAPE and \textit{relpos first layer} have similar mostly uniform performance profiles across different audio durations on RV data. 
This observation asks for reconsidering the standard usage of positional embedding for CTC-based models in speech recognition. 

\subsection{No Positional Embedding Discussion}

As demonstrated above in Figure~\ref{fig:speech:tl}, \textit{nopos} performs similar to different positional embeddings on both public and RV data while having reliable generalization to the long audio fragments.
We figured out that the key components of this phenomenon and \textit{nopos} success are 
i) enough training data; ii) sufficient model capacity and iii) CTC loss.

\begin{figure}[htbp!]
  \centering
  \includegraphics[width=0.9\textwidth]{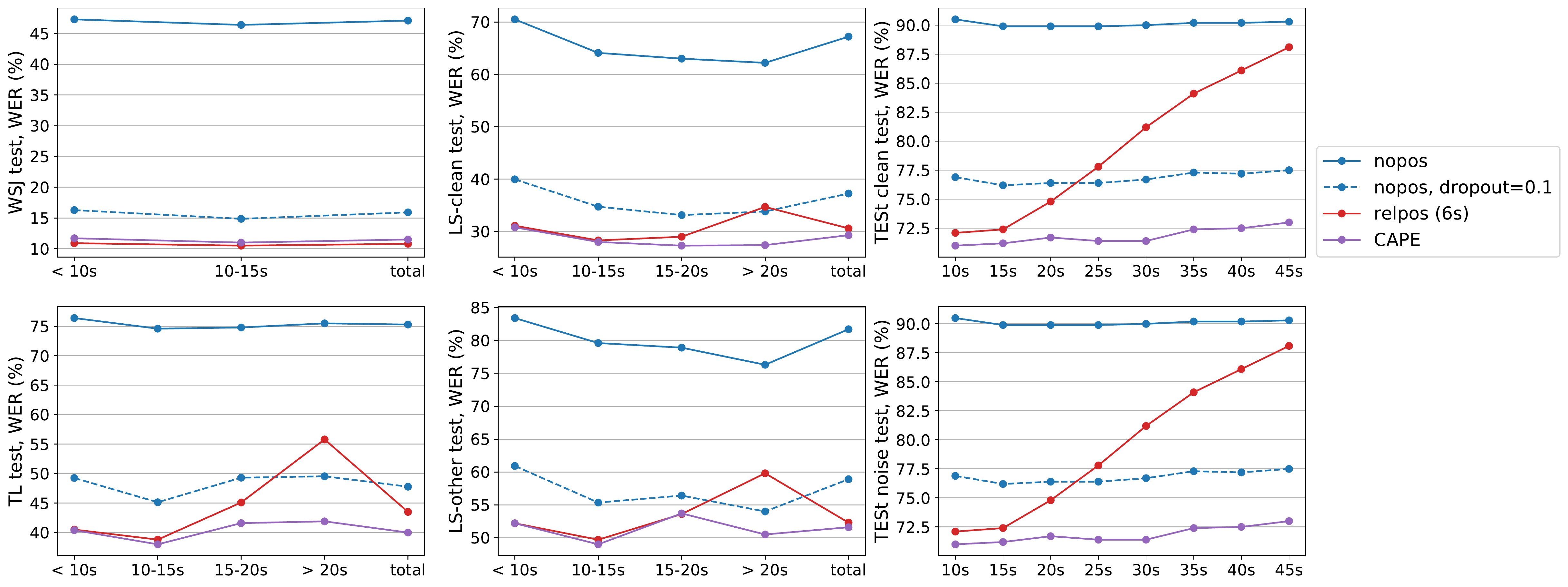}
  \caption{
  Word error rate comparison for models trained on WSJ data with different positional embeddings. 
  Baseline models, \textit{nopos} and CAPE, use 0.4 dropout and 0.4 layer drop in every Transformer layer, while \textit{nopos, dropout=0.1} uses 0.1 for both values.\label{fig:abl:speech:asr_nopos}}
\end{figure}

For the first point we saw that \textit{nopos} model trained on WSJ, a 5x smaller dataset than TL, performs poorly having 45-50\% WER even on in-domain data. 
For the second point we perform an additional ablation on WSJ data by decreasing dropout and layer drop in each Transformer layer from 0.4 to 0.1: with increased model capacity \textit{nopos} reduces the WER by 30\% and gets closer to other positional embeddings, Figure~\ref{fig:abl:speech:asr_nopos}.
For the third point we perform another ablation by comparing with sequence-to-sequence (seq2seq) training: we use exactly the same encoder $\mathbf{H}^{L_e}$ (with various positional embeddings) but replace last linear layer and CTC loss with the decoder, encoder-decoder attention, and cross-entropy loss where the probability distribution of the transcription is factorized as 
\begin{equation*}
p(y_1, ..., y_n) = \prod_{i=1}^n p(y_i \ | \ y_0, ..., y_{i-1}, \mathbf{H}^{L_e})
\end{equation*}
where $y_0$ is a special symbol indicating the beginning of the transcription. 
The decoder is a stack of 6 Transformer layers with encoding dimension 256, learnable relative positional embedding with 9.6s left-only context and 4 attention heads. 
Dropout and layer drop in the decoder layers are set to 0.2.

\begin{figure}[htbp!]
  \centering
  \includegraphics[width=0.85\textwidth]{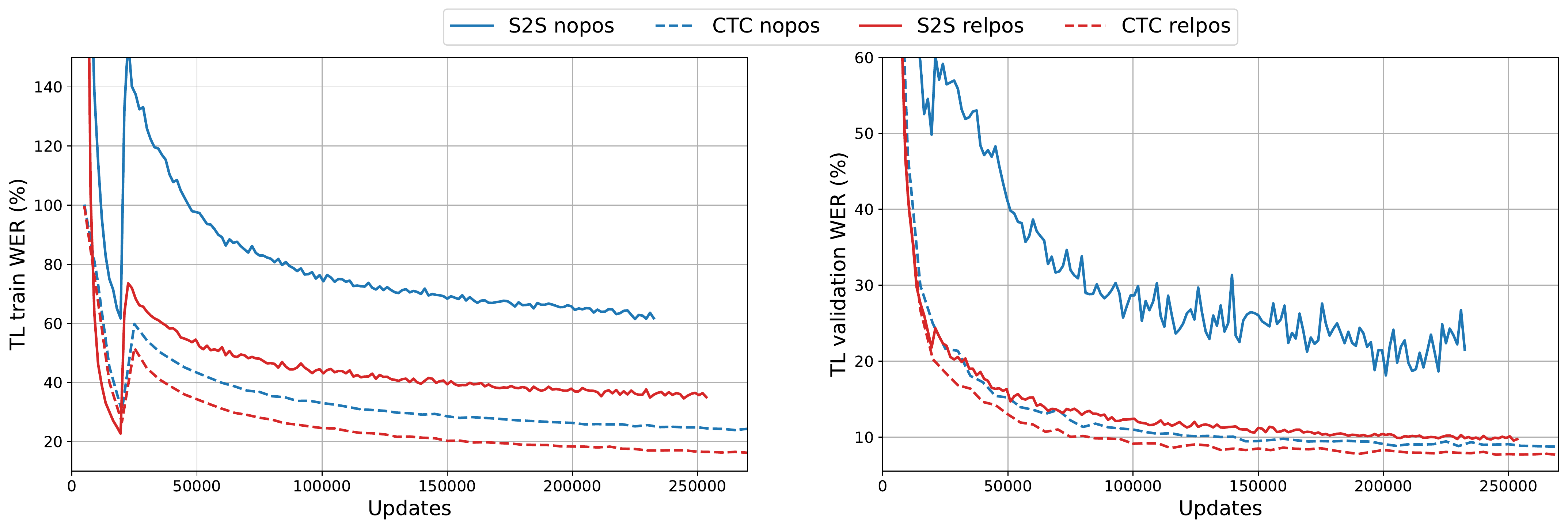}
  \caption{
  Word error comparison for CTC and seq2seq models trained on TED-LIUM v3 data without any positional embedding in the encoder (\textit{nopos}) or with learnable relative positional embedding in every encoder-Transformer layer (\textit{relpos}).\label{fig:abl:speech:asr_s2s}}
\end{figure}

In Figure~\ref{fig:abl:speech:asr_s2s} we show comparison between CTC and seq2seq models trained on TL with either \textit{nopos} or \textit{relpos} in the encoder.\footnote{Encoder remains the same for both CTC and seq2seq models; for seq2seq models decoder is also identical.} 
Seq2seq \textit{nopos} performs significantly worse than seq2seq \textit{relpos} and moreover has higher WER variation on validation data. 
This result is opposite to the \textit{nopos} CTC-based training, suggesting that CTC loss is able to train with enough data or model capacity and no positions provided. 

These observations only partially overlap with known results:
for seq2seq training it was shown recently that relative positions can be modeled via a deep stack of convolutional layers in the encoder~\cite{mohamed2019transformers} or via convolutions inserted directly in each encoder's Transformer layer~\cite{zhang2020pushing}. 
In contrast to the listed works, our encoder has vanilla Transformer layers and only one convolutional layer at the beginning. 
Thus, \textit{nopos} model has very limited context to model relative positions, which affects seq2seq training (it has to ``locate'' corresponding timepoint in audio with attention) more than CTC-based, which uses explicit time ordering.
In line with this interpretation, for the hybrid ASR systems dropping positional information does not drive to significant deterioration of quality~\cite{wang2020transformer}.

\section{Machine Translation}

\subsection{Technical Details}\label{app:mt}

For machine translation experiments we have implemented CAPE within ADMIN's~\cite{liu2019deep,liu2020very} open-sourced code\footnote{\scriptsize \texttt{https://github.com/LiyuanLucasLiu/Transformer-Clinic}} which is based on Fairseq\footnote{\scriptsize  \texttt{https://github.com/pytorch/fairseq}} toolkit~\cite{ott2019fairseq}. 
We precisely follow open-sourced recipes for ADMIN with \textit{sinpos}\footnote{\scriptsize  \texttt{https://github.com/LiyuanLucasLiu/Transformer-Clinic/blob/master/nmt-experiments}: \texttt{wmt14\_en-de.md} and \texttt{wmt14\_en-fr.md}} including data preparation step; the only change we introduce is usage of different positional embeddings. Besides, we do not perform model averaging as was done in~\cite{liu2019deep,liu2020very}. 

All English-German (DE) models are trained for 150 epochs on 4 GPUs (Volta V100 16GB) for 30h (6l-6L) or 70h (18L-18L). 
English-French (FR) 6L-6L models are trained for 75 epochs on 8 GPUs (Volta V100 16GB) for 60h. 
Each configuration is trained starting from 3 different random seeds.

As mentioned in Section~\ref{sec:mt} we scale positions of source language by a factor $\alpha$ computed based on train data statistics only as ${\alpha = \frac{\text{\# tokens in target corpus}}{\text{\# tokens in source corpus}} \in\mathbb{R}}$: it is set to $\alpha=1.0337$ for DE and $\alpha=1.1632$ for FR. 
For all machine translation experiments with CAPE we skip the mean-normalization step to have source and target sentences aligned at the first position and prepend ``begin of sentence'' in the source sentences to give a hint of first position for the model's decoder (as after global shift there is no way to determine the first position from its positional embedding anymore).
Additionally, we do not apply any global scaling. 
For the global shift we sweep values 5, 10 and 50 while the local shift is set to maximum to preserve positions order, $\epsilon_{max}=0.5$. 

\begin{table}[h!]
\caption{Slowdown for relpos / no relpos. (Top) 1000 context left/right, (Bottom) 100 context left/right. Run on V-100 GB32, 100 runs, 10 runs warmup, batch 50, emb 768, head 8. \label{tab:speed_relpos}}
\begin{center}
\resizebox{0.9\linewidth}{!}{
\begin{tabular}{@{}ccccccc@{}}
\toprule
Model & FP16 Len-10 & FP32 Len-10 & FP16 Len-100 & FP32 Len-100 & FP16 Len-1000 & FP32 Len-1000 \\
\midrule
Transformer layer & 2.1 & 2.3 & 3.3 & 2.3 & 2.2 & 1.7 \\
\midrule
Encoder & 2.2 & 2.2 & 2.6 & 1.7 & 2.0 & 1.6 \\
\midrule
Decoder & 1.8 & 1.7 & 1.9 & 1.4 & 1.6 & 1.4 \\
\midrule
\midrule
Transformer layer & 2.1 & 2.1 & 1.9 & 1.2 & 1.4 & 1.2 \\
\midrule
Encoder & 2.3 & 2.4 & 1.5 & 1.1 & 1.4 & 1.2 \\
\midrule
Decoder & 1.7 & 1.7 & 1.2 & 1.1 & 1.2 & 1.1 \\
\bottomrule
\end{tabular}
}
\end{center}
\end{table}

\subsection{Computational Cost}

For machine translation experiments we estimate the forward and backward time slowdown when learnable relative positional embedding is used in Transformer layer compared to vanilla Transformer layer. 
Besides we evaluate slowdown for entire encoder and decoder. 
Forward and backward time benchmarking is done in Fairseq (PyTorch) for different input sequences length (10, 100, 1000), for different context size of relative positional embedding (100 and 1000 tokens to left/right) in full-precision (fp32) and half-precision (fp16) floating-point computations.
We use Volta V-100 32GB GPU and report slowdown for average time of forward and backward passes measured across 100 updates with 10 updates of warmup, see Table~\ref{tab:speed_relpos}. The batch size is set to 50, attention heads are set to 8 and embedding dimension is 768 (typical use case), there are 6 Transformer layers in encoder and decoder. Results in Table~\ref{tab:speed_relpos} demonstrate that relative positional embedding indeed has additional computational cost which is even more pronounced with mixed-precision computations.

\end{document}